%% file: acl_latex.tex
\documentclass[11pt]{article}

\usepackage[preprint]{acl}

\usepackage{times}
\usepackage{latexsym}

\usepackage[T1]{fontenc}

\usepackage[utf8]{inputenc}

\usepackage{microtype}

\usepackage{inconsolata}

\usepackage{graphicx}

\usepackage{booktabs}
\usepackage{wrapfig}
\usepackage{picinpar}

\usepackage{enumitem}
\usepackage{amsmath,amssymb}

\usepackage{cleveref}

    \usepackage{ai-usage-card}
\usepackage{annotation}
    \aiProjectName{Is Visual Prompting All You Need?\\ Studying VLM Spatial Reasoning under Progressive Visual Scaffolds}
    \aiDomain{Multimodal Machine Learning}
    \aiKeyApplication{Spatial Visual Reasoning}
    \aiContactName{Lars Benedikt Kaesberg}
    \aiContactEmail{l.kaesberg@uni-goettingen.de}
    \aiContactAffiliation{University of Göttingen, Germany}
    \aiModels{Claude Opus 4.7}

    \aiFindingLiterature{Claude}
    \aiFindingExamples{Claude}

    \aiImprovingContent{Claude}

    \aiGeneratingCode{Claude}
    \aiRefactoringCode{Claude}

    \aiWhyUse{Speed, finding new perspectives, proofreading}
    \aiMitigateErrors{Code Reviews}

\title{Is Visual Prompting All You Need?\\ Studying VLM Spatial Reasoning under Progressive Visual Scaffolds}

\author{
  Lars Benedikt Kaesberg$^{*}$, Tianyu Yang$^{*}$, Florian Valentin Wunderlich, Terry Ruas,\\
  \textbf{Daniel Kurzawe, Jan Philip Wahle$^{\dagger}$, Bela Gipp$^{\dagger}$}\\
  University of Göttingen, Germany\\[0.6em]
  \normalsize\textbf{\textsuperscript{*}Equal contribution}\quad
  \textbf{\textsuperscript{\textdagger}Shared last authorship}\\
  \normalsize\textbf{Correspondence:} \texttt{\{l.kaesberg, tianyu.yang\}@uni-goettingen.de}\\[1em]
}

\begin{document}
\maketitle
\AddAnnotationRef{}
\begin{abstract}
Vision-language models (VLMs) have advanced rapidly in multimodal reasoning, yet recent work shows that their failures often reflect an interaction between visual grounding and downstream reasoning.
What remains less clear is how the visual presentation of a task shapes model performance and failure modes when the underlying reasoning problem is unchanged.
We study this question in SPaRC, a benchmark for grid-based visual spatial planning, by introducing lightweight input-side scaffolds that preserve the visual modality while making spatial structure more accessible.
The scaffolds are task-specific diagnostic interventions built on SPaRC's known board structure, not a general-purpose prompting strategy.
Across multiple VLMs, these scaffolds improve task accuracy over the original visual setting by up to 34.0 percentage points and further complement GRPO-based training, yielding up to 4.6 additional accuracy points compared with near-zero gains on the original visual input.
Analyses on both end-to-end task solving and object detection show that these gains are closely tied to reductions in grounding-related errors, while rule reasoning remains comparatively challenging.
With the planning problem held fixed, visual presentation therefore changes both accuracy and the distribution of failure modes, which in turn determines whether a VLM benchmark measures grounded perception, downstream reasoning, or a mixture of both.

\end{abstract}

\section{Introduction}

Vision-language models (VLMs) have rapidly advanced as a general framework for integrating visual perception with language-based reasoning~\cite{singh2025openai, comanici2025gemini, bai2025qwen3, wang2025internvl3}. 
By augmenting large language models (LLMs) with pretrained visual encoders such as CLIP-ViT~\cite{NEURIPS2023_6dcf277e, radford2021learning}, VLMs integrate visual inputs into language modeling, enabling them to support increasingly information-intensive tasks and establishing them as a key building block for applications ranging from embodied artificial intelligence (AI) to world models~\cite{10.1093/nsr/nwae403}.
Yet, despite rapid progress, the mechanisms underlying the multimodal understanding ability of VLMs remain poorly understood.

\begin{figure}
    \centering
    \includegraphics[width=\columnwidth]{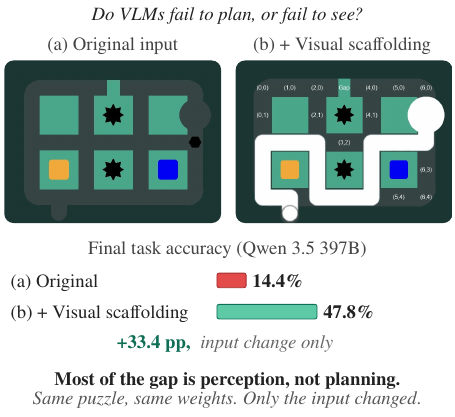}
    \caption{Impact of visual scaffolding on task solving.}
    \label{fig:figure-abstract}
\end{figure}

Recent work shows that apparent multimodal reasoning failures often reflect how visual information is grounded, verbalized, or bypassed, not just downstream reasoning limits~\cite{wu2025vsp,asadi2026mirage,shahgir2026vlms}.
Prior work addresses such failures through model-side adaptation~\cite{chen2025bring, zhang2026viper, jain2026elevating,DBLP:journals/corr/abs-2509-13031}, inference-time visual tools~\cite{hong2026deepeyesv}, or input-side image scaffolding~\cite{yang2023setofmark}, yet none isolate which dimensions of the visual representation itself induce difficulty for planning.
Diagnostic studies often replace images with text or probe isolated subtasks~\cite{kaesberg-etal-2025-sparc, wu2025vsp}, sidestepping the visual interface or obscuring its effects on downstream planning.
We therefore ask how VLM task performance and failure modes change as the same planning problem is presented with progressively stronger visual support.

SPaRC~\cite{kaesberg-etal-2025-sparc} is a grid-based visual puzzle benchmark that provides a controllable testbed for studying this question. 
Each puzzle presents a board with structured elements such as grid cells, rule symbols, and start/end positions, and requires the model to construct a valid path under explicit constraints. 
Because solving the task depends on both perceiving these elements and reasoning over their relations, errors in visual grounding can easily propagate into downstream planning failures, making the source of model weakness difficult to isolate in the standard evaluation setting. 
In this paper, we therefore construct a hierarchy of lightweight visual scaffolds, ranging from the original puzzle image to variants with explicit start/end markers, coordinate grids, cell-level coordinate annotations, and text-rendered symbols. 
Using this hierarchy as a diagnostic intervention, we evaluate end-to-end task accuracy, object detection accuracy, and path-level error categories covering endpoint grounding, path connectivity, self-intersection, and rule satisfaction. 
Together, these evaluations allow us to track how model performance and failure modes change as perceptual support increases, while the underlying board layout, rule configuration, and target solution remain fixed.

Our results reveal a consistent pattern across the tested configurations.
Without any model fine-tuning, increasing input-side perceptual support substantially improves VLM performance on SPaRC, with gains of up to 34.0 percentage points over the original visual format.
These large zero-shot gains show that the input representation itself accounts for a substantial share of the difficulty measured on this benchmark, even for strong frontier-scale models.
When we fine-tune models using GRPO~\cite{shao2024deepseekmath}, visual interventions yield larger training gains under scaffolded visual inputs than under either the original visual input or the text-formulation baseline.
These improvements are reflected in both aggregate accuracy and the structure of model errors. Endpoint localization, coordinate grounding, and invalid path construction have improved substantially, while rule satisfaction remains comparatively challenging.
These findings refine prior evidence on perception bottlenecks: rather than showing that perception is \emph{the} bottleneck, they show that with the planning problem held fixed, visual presentation changes both accuracy and the distribution of failure modes.

Overall, our contribution can be summarized as:
\begin{itemize}[leftmargin=*,label={\color{teal}$\blacktriangleright$},itemsep=0.15em]
    \item We revisit multimodal reasoning from a data-centric perspective and argue that input-side perceptual support can materially affect downstream planning performance. 
    \item Using the visual spatial planning benchmark SPaRC as a testbed, we introduce a hierarchy of lightweight visual scaffolds that progressively increase perceptual support and use it to diagnose VLM behavior on both the planning task and a board-element detection task.

    \item Extensive experiments over the benchmark show that our scaffolding hierarchy separates grounding errors from residual reasoning challenges, suggesting that visual presentation shapes not only model performance, but also how perception and reasoning are entangled in benchmark evaluation.\footnote{Our code and data are available on \href{https://github.com/flowun/sparcVisualPuzzleEvaluation}{GitHub}.}
\end{itemize}

\section{Related Work}
\subsection{Multimodal Reasoning in VLMs}

VLMs extend LLMs with pretrained visual encoders and achieve strong performance across visual question answering, chart and document understanding, embodied decision making, and multimodal agent tasks~\cite{NEURIPS2023_6dcf277e, team2026kimi, bai2025qwen3, wang2025internvl3, zheng2026deepeyes, yang2025alden, hy2026hy, team2026qwen3}.
End-to-end accuracy on such benchmarks, however, conflates the ability to perceive the visual input with the ability to reason over what was perceived~\cite{wang2026perceptionaware}.
This conflation is especially acute in visual spatial planning, where a single failure mode may reflect either a reasoning limitation or an earlier failure to ground the spatial structure the reasoning depends on.
Our work directly targets this ambiguity by varying only the visual representation while holding the underlying planning task fixed.

\subsection{Visual Spatial Planning and Representation Design}

In visual spatial planning, image-based inputs consistently underperform textual descriptions of the same environment, a gap commonly read as a fundamental modality limitation~\cite{wu2025vsp, aghzal2024can}.
For instance, o4-mini drops from $15.8\%$ accuracy on the SPaRC text version to $5.6\%$ on the visual input
\citep{kaesberg-etal-2025-sparc}.
Prior mitigations adapt the model via fine-tuning or merging~\cite{chen2025bring, zhang2026viper, jain2026elevating, DBLP:journals/corr/abs-2509-13031}, add inference-time visual tools such as crop or refocus~\cite{yang2026look, guo2025beyond, wu2026reinforcing}, or scaffold the image with overlays, sketches, or structured
renderings~\cite{yang2023setofmark, hu2024visual, menon-etal-2024-whiteboard}.
Our proposal is most similar to the last one, but it uses single scaffolds for grounding instead of isolating which factors of perception drive reasoning.
CoordConv makes a related point outside VLMs: convolutional networks fail trivial coordinate-transform tasks until $(i,j)$ coordinate channels are added to the input~\citep{NEURIPS2018_60106888}.
Our Axis Labels and Cell Coordinates scaffolds render coordinates into the image instead. We propose no architecture, and unlike that task-agnostic module our scaffolds are built from SPaRC's known structure (\S\ref{sec:3.2}).
We intervene on the input representation itself by progressively making endpoint identity, spatial indexing, and symbol semantics explicit while preserving the planning task, which allows us to identify which factors close the vision-text gap and whether the gap is driven by modality or representation.

\begin{figure*}[t]
\begin{center}
\includegraphics[width=1.0\textwidth]{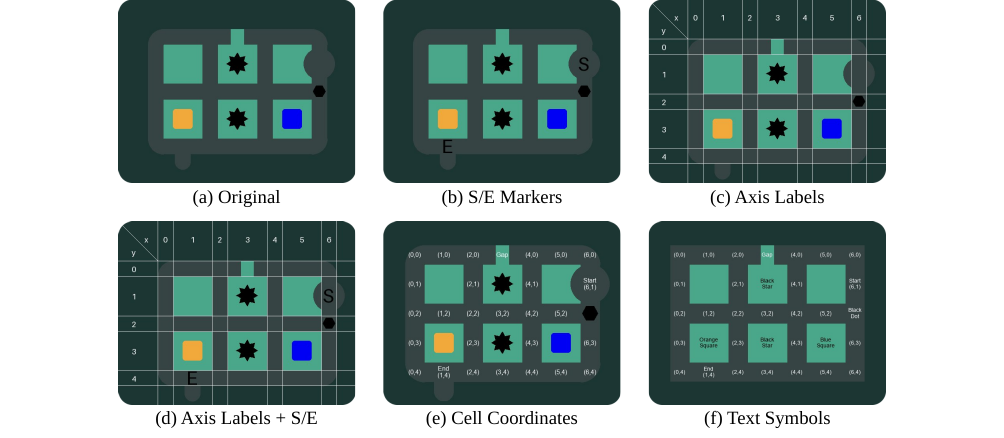}
\end{center}
\caption{Overview of the input representations used in our study: \textbf{Original}, \textbf{S/E Markers}, \textbf{Axis Labels}, \textbf{Axis Labels + S/E}, \textbf{Cell Coordinates}, and \textbf{Text Symbols}. These variants progressively increase perceptual support by making endpoint identity, spatial indexing, and symbol semantics more explicit, while keeping the underlying planning task unchanged.}

\label{figure:1}
\end{figure*}
\section{Methodology}
In the following, we describe the methodology of this paper, including the task setting (\S \ref{sec:3.1}), input representations (\S \ref{sec:3.2}), input conditions (\S \ref{sec:3.3}), and evaluation metrics (\S \ref{sec:3.4}).

\subsection{SPaRC as a Testbed}
\label{sec:3.1}
We conduct our study on SPaRC \cite{kaesberg-etal-2025-sparc}, a benchmark for grid-based visual spatial planning.
The task takes place on a 2D grid, and each sample is defined by a grid $G$ of dimensions $m \times n$. 
This grid is populated with a set of rule cells $R$, a designated start node $S$, and a designated end node $E$. 
The dataset has 1000 examples (500 for training and 500 for testing) with different sizes and difficulty levels, ranging from level 1 (easiest) to level 5 (hardest).
The objective for the agent is to generate a sequence of coordinates representing a valid continuous path
\[
C=(c_0,c_1,\dots,c_k),
\]
where each coordinate is denoted as \(c_i=(x_i,y_i)\).
See Appendix~\ref{app:sparc_dataset} for the full rule set and further details on the dataset.

\subsection{Input-Side Visual Scaffolding}
\label{sec:3.2}
To study how perceptual accessibility affects spatial planning, we construct a hierarchy of board representations that progressively expose spatial structure while preserving the underlying task.
Each step adds a specific form of perceptual support, and we use changes in downstream error profiles as diagnostic measurements. 
Figure~\ref{figure:1} provides examples of each representation, and we describe the added support and the error types expected to be most sensitive to it.

\textbf{S/E Markers (S/E):} Retains the original board layout and symbols, but explicitly marks the start (\texttt{S}) and exit (\texttt{E}) cells, providing the lowest level of perceptual assistance.
We expect this to be most relevant to endpoint-localization errors and downstream complete-path failures from incorrect start or exit grounding.

\textbf{Axis Labels (Axis):} Adds row and column indices along the board borders, analogous to spreadsheet-style coordinates, providing global spatial-reference support. 
We expect it to affect coordinate-referencing and spatial-alignment errors, as well as path-structure errors from unstable global grounding.

\textbf{Axis Labels + S/E (Axis+S/E):} Combines \textbf{Axis} and \textbf{S/E}. We expect it to reduce endpoint- and coordinate-related errors, with possible downstream benefits for path connectivity and complete-path validity.

\textbf{Cell Coordinates (Cell Coord.):} Extends spatial scaffolding by printing each cell's coordinate (e.g., $(1,3)$) directly inside the board, removing ambiguity in cell localization and providing stronger local spatial-grounding support. 
We expect this to be especially relevant to localization, coordinate-binding, path-connectivity, and complete-path-validity errors.

\textbf{Text Symbols (Text Sym.):} Builds on \textbf{Cell Coord.} by replacing graphical rule symbols with text labels (e.g., a star icon as the word ``star''), making both spatial indexing and symbol identity explicit.
We expect it to be most relevant to symbol-recognition and rule-grounding errors, while also revealing residual constraint-reasoning errors that persist when visual information is explicit.

Figure~\ref{figure:1} illustrates this progression from the original visual board to increasingly scaffolded variants, ending with a representation that renders rule symbols as text.
Across all variants, we keep the board layout, rule configuration, and planning objective fixed, changing only how spatial information is presented.
Later representations inherit support from earlier ones and may affect multiple error types simultaneously, so we do not assume a one-to-one correspondence between representation and failure category. 
We use the hierarchy to examine how the model's error profile
changes as perceptual burden is reduced, and in particular where
grounding-sensitive failures give way to residual reasoning
failures. The original SPaRC textual formulation is reported
separately in Appendix~\ref{app:text-baseline} as a non-visual
reference.

\paragraph{Scope.}
Each scaffold is built from SPaRC's known structure: the task requires identifying the start and exit, emitting coordinates, and reading rule symbols, and each variant makes one of these explicit.
They are therefore task-specific diagnostic interventions rather than a general prompting strategy.
The steps are also cumulative, since later variants add coordinate and semantic cues along with visual structure; Appendix~\ref{app:grid_label_ablation} decomposes the Axis Labels step. Text Symbols in particular shifts part of the grounding burden toward reading in-image text.

\subsection{Visual Degradation and Recovery}
\label{sec:3.3}
To probe how much of the observed difficulty is tied to visual grounding, we construct three settings that reduce the quality of the original board image while preserving the underlying planning problem (see Appendix~\ref{app:recovery_variants} for examples). 
\textbf{Low Contrast} desaturates the board colors, making symbols less visually distinctive. 
\textbf{Low Resolution} downsamples the image, reducing fine-grained pixel detail. 
\textbf{Rotated} rotates the entire board, disrupting its canonical orientation. 
For each degradation, we construct a recovery variant by overlaying the \textbf{Cell Coordinates} representation on the degraded board, allowing us to test whether degraded performance can be recovered by restoring access to the board's spatial structure, without also converting symbol identities into text.

\subsection{Evaluation Tasks and Metrics}
\label{sec:3.4}

We evaluate model behavior under two complementary settings: end-to-end task solving and object detection, which together let us analyze how the representation hierarchy in \S\ref{sec:3.2} affects both planning behavior and the visual grounding that supports it.

\paragraph{Task solving.}
The model is given a board representation and asked to generate a path from the start cell to the exit that satisfies the rule constraints. Our primary metric is \textbf{Final Task Accuracy}, the fraction of puzzles for which the predicted path is completely correct. To track how failures shift with visual scaffolding, we additionally define four path-level validity conditions. \textbf{Endpoint correctness} checks whether the generated path connects the designated start and exit cells.
\textbf{Connectivity} requires every consecutive pair of coordinates to be adjacent under Manhattan distance, i.e., $\|c_{i+1}-c_i\|_1 = 1$. \textbf{Non-intersection} requires that the path not revisit any coordinate.
\textbf{Rule-cell avoidance} requires that the path never enter a rule cell, i.e., $c_i \notin \mathcal{R}$ for all $i$. We report the satisfaction rates of these four conditions as \textbf{Correct Start/End}, \textbf{Connected Path}, \textbf{Non-Intersecting}, and \textbf{No Rule Violation}, and additionally report \textbf{Fully Valid}, the fraction of paths satisfying all four at once.

\paragraph{Object detection.}
The model is asked to identify the rule symbols present on the board, providing a direct readout of visual recognition that complements the path-level analysis. We report \textbf{Exact-Match Board Accuracy} (the fraction of boards on which all rule symbols are identified correctly), \textbf{Average Rule Detection Accuracy} (per-rule accuracy averaged across boards), and \textbf{Per-Type Detection Accuracy} (accuracy per symbol type). Comparing Cell Coordinates with Text Symbols isolates gains from local spatial grounding versus explicit symbol semantics.

Detection uses a separate prompt (Appendix~\ref{app:prompt-detection}) and is independent of the planning answer. The model sees the board image and emits the whole board as a 2D array of symbol codes in a fixed schema, which a deterministic script parses and compares cell by cell against the ground truth. For a ground-truth Start cell, a predicted label of Path, End, or Gap contributes to the corresponding off-diagonal entry of the confusion matrices in \S\ref{sec:4.3}.

\begin{figure*}[t]
\begin{center}
\includegraphics[width=0.9\textwidth]{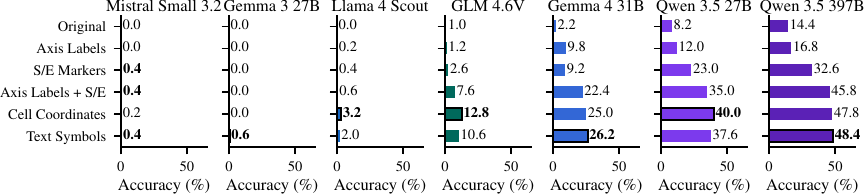}
\end{center}

\caption{Task-solving accuracy across board representations. We compare the \textbf{Original} SPaRC visual input and progressively scaffolded visual variants. The best results are highlighted in \textbf{bold}.
}
\label{fig:board_comparison}
\end{figure*}

\section{Results and Analysis}

In this section, we present our experimental findings. 
We first describe the experimental setup (\S\ref{sec:4.1}), then analyze zero-shot behavior through end-to-end planning accuracy (\S\ref{sec:4.2}), board-element detection (\S\ref{sec:4.3}), and path-level error profiles (\S\ref{sec:4.4}). 
We further evaluate whether scaffolding recovers performance under visual degradation (\S\ref{sec:4.5}) and whether it complements GRPO post-training (\S\ref{sec:4.6}). 

\begin{figure}[t]
    \centering
\includegraphics[width=0.9\columnwidth]{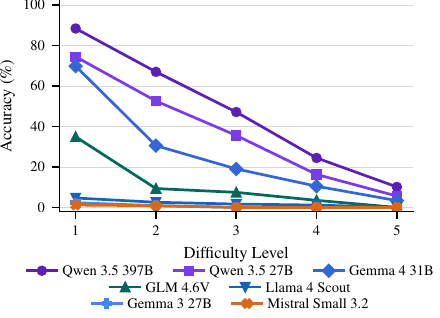}
    \caption{Task-solving accuracy stratified by puzzle difficulty for Text Symbols representation. 
    }
    \vspace{-12pt}
    \label{fig:difficulty_comparison}
\end{figure}

\subsection{Experimental Setup}
\label{sec:4.1}
We evaluate seven open-weight VLMs, namely Mistral Small 3.2~\cite{mistralsmall32}, Gemma 3 27B~\cite{gemmateam2025gemma3technicalreport}, Llama 4 Scout~\cite{llama4}, GLM 4.6V~\cite{vteam2026glm45vglm41vthinkingversatilemultimodal}, Gemma 4 31B~\cite{gemma4}, Qwen 3.5 27B and 397B~\cite{qwen3.5}. The selection covers dense and Mixture-of-Experts architectures with 24B to 397B parameters. 
For the GRPO experiments (\S \ref{sec:4.6}), we use Qwen 3 VL 4B Thinking and 8B Thinking~\cite{bai2025qwen3} as base models. 
All inferences use greedy decoding unless stated otherwise. 
Technical details about the models and hardware, including the configuration used for GRPO training, are reported in \Cref{app:models_hardware}.

The test split contains 500 puzzles. Appendix~\ref{app:bootstrap} reports $95\%$ bootstrap confidence intervals for every model and condition, and paired intervals for the differences we discuss.

\begin{figure*}[t]
\begin{center}
\includegraphics[width=0.9\textwidth]{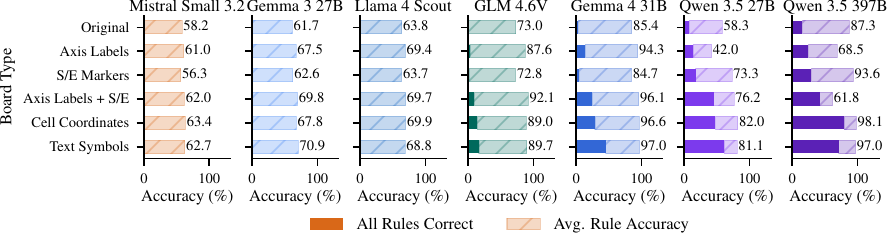}
\end{center}
\caption{Object-detection performance across board representations. 
We report exact-match board accuracy and average rule detection accuracy for each model and representation. }
\label{fig:rule_accuracy_comparison}
\end{figure*}

\begin{figure}[t]
    \centering
\includegraphics[width=0.9\columnwidth]{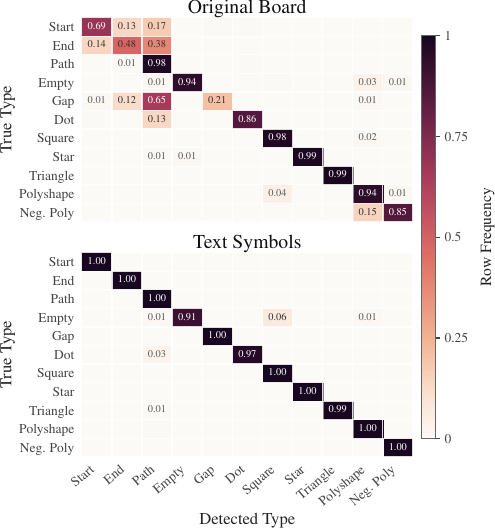}
    \caption{Per-type detection accuracy and confusion patterns for the original board and Text Symbols representation with Qwen 3.5 397B.}
    \label{fig:od_confusion_comparison}
\end{figure}

\subsection{Effect of Visual Scaffolding on Task Solving}
\label{sec:4.2}

We examine end-to-end task-solving performance across the representation hierarchy introduced in \S \ref{sec:3.2}. 
As shown in Figure~\ref{fig:board_comparison}, input-side visual scaffolding consistently improves accuracy over the original visual board, especially for stronger models. 
For example, Qwen~3.5~397B improves from 14.4\% on the Original board to 48.4\% with Text Symbols, Qwen~3.5~27B improves from 8.2\% to 40.0\% with Cell Coordinates, and Gemma~4~31B improves from 2.2\% to 26.2\% with Text Symbols. Paired bootstrap intervals exclude zero for all three (Appendix~\ref{app:bootstrap}).
The two models at the accuracy floor are the exception: Gemma~3~27B and Mistral~Small~3.2 stay below $1\%$ on every visual condition and show no resolvable gain from any scaffold.
These gains suggest that a large part of the difficulty measured on the Original board is attributable to perceptual access rather than to planning difficulty alone, and that perceptual support is most useful when models have enough planning capacity to exploit it, a pattern further supported by the auxiliary completion-token analysis in Appendix~\ref{app:token_usage}.

The best scaffolded visual inputs exceed the SPaRC text-formulation baseline. Qwen~3.5~397B reaches $48.4\%$ with Text Symbols versus $38.0\%$ on the text baseline, a \textbf{+10.4 pp} gain from changing the input format alone. The text formulation is thus not an upper bound on reasoning ability for this task. Per-model text-vs-visual comparisons and the human reference are in Appendix~\ref{app:text-baseline}.

Finally, Figure~\ref{fig:difficulty_comparison} shows task-solving accuracy by puzzle difficulty under the Text Symbols representation. 
Accuracy decreases monotonically for all models as difficulty increases, indicating that perceptual support does not eliminate the need for nontrivial planning. 
The strongest models benefit most on easier and medium-difficulty instances, while performance drops sharply at levels 4--5 and the gap between models narrows. 
This suggests that scaffolding helps when the remaining planning problem is within the model's capacity, but harder puzzles still expose residual long-horizon planning and rule-reasoning limitations.

\subsection{Effect of Visual Scaffolding on Perceptual Recognition}
\label{sec:4.3}

Next, we examine whether improvements in task-solving accuracy are accompanied by stronger perceptual recognition by asking models to identify the rule symbols present on the board. 
Figure~\ref{fig:rule_accuracy_comparison} reports object-detection performance across board representations. 
Overall, scaffolded inputs improve rule recognition over the original visual board for most models. 
The effect is especially clear for stronger models.
Gemma 4 31B improves from 85.4\% average rule accuracy on the original board to 97.0\% with Text Symbols, while Qwen 3.5 397B improves from 87.3\% to 98.1\% with Cell Coordinates. 
GLM 4.6V also shows improvements under coordinate-based variants, reaching 92.1\% with Axis Labels + S/E.

The gains are not uniformly monotonic across all models and representations.
For example, Qwen 3.5 27B shows lower detection accuracy with Axis Labels than with the original board, while Qwen 3.5 397B performs worse under Axis Labels + S/E than under the original board. 
This suggests that adding visual scaffolds can introduce clutter or change the visual distribution in ways that affect recognition. 
Nevertheless, the trend shows that representations with stronger spatial support improve perceptual access to board elements.

Figure~\ref{fig:od_confusion_comparison} provides a finer-grained view through per-type confusion patterns for Qwen~3.5~397B. 
On the Original board, several structural elements are difficult to distinguish. 
Start and End are correctly detected only \(69\%\) and \(48\%\) of the time, respectively, with \(17\%\) of Start cells and \(38\%\) of End cells misclassified as Path. 
Gap is the most severe case: only \(21\%\) of Gap cells are correctly detected, while \(65\%\) are misclassified as Path. 
Text Symbols sharply reduces these confusions. 
Start, End, Path, and Gap all reach \(100\%\) detection accuracy, and most rule symbols also become nearly perfect, including Square, Star, Polyshape, and Neg. Poly at \(100\%\), Triangle at \(99\%\), and Dot at \(97\%\). 
This shift supports the view that explicit symbolic rendering reduces ambiguity in the visual input and improves grounding of board components.
This confusion pattern is not specific to Qwen~3.5~397B. 
An averaged confusion analysis across all seven models, reported in Appendix~\ref{app:avg_confusion}, shows the same collapse of symbol classes onto the dominant Path class on the Original board and the same diagonal sharpening under Text Symbols.

\begin{figure}[t]
    \centering
\includegraphics[width=0.9\columnwidth]{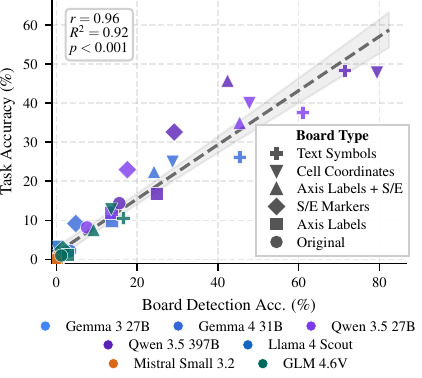}
    \caption{Correlation between object-detection performance and end-to-end task-solving accuracy across models and board representations. }
    \label{fig:od_task_correlation}
\end{figure}

Finally, Figure~\ref{fig:od_task_correlation} compares object-detection performance with end-to-end task-solving accuracy across models and board representations. 
The strong positive correlation (\(r=0.96\), \(R^2=0.92\), \(p<0.001\)) indicates that representations and models with better perceptual recognition also tend to achieve higher planning accuracy. 
This should not be read as proving that perception alone determines task success, since both detection and planning may depend on model capacity. 
Together with the hierarchy results in \S\ref{sec:4.2}, it indicates that improved perceptual grounding is an important mediator of the scaffolding gains.
Appendix~\ref{app:avg_correlation} repeats this correlation with Average Rule Detection Accuracy, where it remains significant but weaker $(r=0.51)$.

\subsection{Path-Level Error Analysis}
\label{sec:4.4}

\begin{figure}[t]
    \centering
\includegraphics[width=0.9\columnwidth]{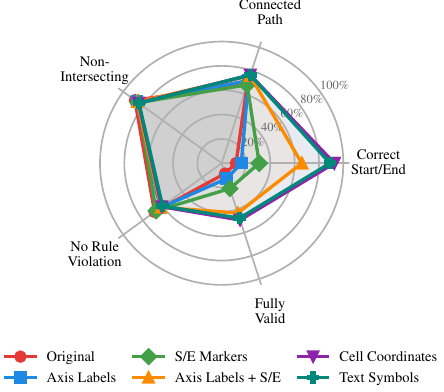}
    \caption{Path-level validity analysis for four representative input settings averaged across all models. }
    \label{fig:path_analysis_radar}
\end{figure}

To understand how visual scaffolding affects model behavior, we analyze the generated paths using the validity metrics from \S\ref{sec:3.4}.
Figure~\ref{fig:path_analysis_radar} shows the results, averaged across all evaluated models.

The metrics most sensitive to scaffolding are \textbf{Correct Start/End}, \textbf{Connected Path}, and \textbf{Fully Valid}.
On the Original board, only 11.8\% of paths connect the right endpoints and 8.9\% are Fully Valid.
Cell Coordinates and Text Symbols push Correct Start/End to around 90\% and Fully Valid to roughly 48\%, while Connected Path moves more gently (68.2\% to 76.3\%).
For comparison, the SPaRC text baseline reaches 97.2\% endpoint correctness, 85.1\% connectivity, and 46.7\% Fully Valid, so the two strongest scaffolds reach comparable Fully Valid rates.
This is a controlled reduction of the visual grounding burden, not a solution to the original visual task: Text Symbols shifts part of that burden toward reading in-image text, though it still preserves the spatial layout that the text formulation linearizes.
Under the Original visual format, the dominant failure mode is unstable spatial grounding rather than planning.

\textbf{Non-Intersecting} stays high across all configurations (84.0--88.1\%, against 89.8\% on text) and barely shifts, so revisiting cells is not a meaningful failure mode.
\textbf{No Rule Violation}, which checks whether the path stays off rule cells, sits between 60.0\% and 68.1\% across the visual variants and 52.9\% on the text baseline, with no clear trend across the scaffolding hierarchy.
The Fully Valid gains under scaffolding are therefore driven by improvements in endpoint and connectivity and not by changes in these two conditions.

\subsection{Recovery from Visual Degradation}
\label{sec:4.5}

We next use the degradation-and-recovery setting from \S\ref{sec:3.3} as a stress test of the grounding hypothesis. 
If scaffolding restores access to the board's spatial structure, degraded inputs should reduce performance, while the corresponding Cell Coordinates variants should remain closer to their clean counterpart. 

\begin{figure}[t]
    \centering
\includegraphics[width=0.9\columnwidth]{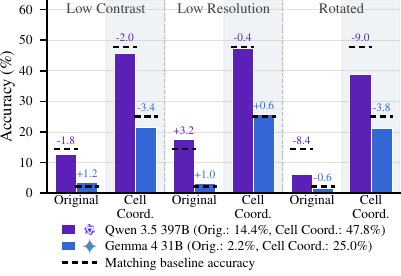}
    \caption{Task-solving accuracy under visual degradation and scaffolded recovery settings.}
    \label{fig:worsening_comparison}
\end{figure}

Figure~\ref{fig:worsening_comparison} reports changes in task-solving accuracy relative to the clean Original and Cell Coordinates settings.
Low Contrast and Low Resolution have limited effects under Cell Coordinates. 
For Qwen~3.5~397B, the recovered variants change by only \(-2.0\) and \(-0.4\) points, respectively; for Gemma~4~31B, the corresponding changes are \(-3.4\) and \(+0.6\) points. 
Rotation is more disruptive, with larger drops under Rotated + Cell Coordinates (\(-9.0\) for Qwen~3.5~397B and \(-3.8\) for Gemma~4~31B), suggesting that orientation changes introduce an additional spatial-alignment challenge.

A complementary analysis of object-detection accuracy under the same degradations (Appendix~\ref{app:od_degradation}) shows a similar pattern. 
Rule detection remains largely stable under Low Contrast and Low Resolution, but drops more clearly under Rotation, especially for the Cell Coordinates variants. 
Overall, Cell Coordinates preserve much of the performance under degradations that reduce salience or resolution, while rotation remains a harder case. 
This supports the view that explicit spatial scaffolding can recover impaired visual grounding, but does not fully eliminate orientation-sensitive spatial alignment errors.

\subsection{Effect on GRPO Post-Training}
\label{sec:4.6}

\begin{figure}
    \centering
    \includegraphics[width=0.9\linewidth]{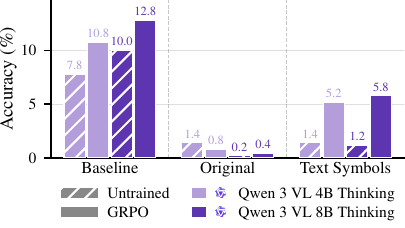}
    \caption{Effect of GRPO post-training on task-solving accuracy for Qwen 3 VL 4B and 8B Thinking, evaluated across three input representations.}
    \label{fig:grpo-training}
\end{figure}

To test how the input representation interacts with post-training, we apply GRPO to Qwen 3 VL 4B Thinking and 8B Thinking under three input conditions: the text-only \textbf{Baseline}, the unannotated \textbf{Original} image, and the fully scaffolded \textbf{Text Symbols} variant. 
More details about the setting can be seen in Appendix~ \ref{app:grpo-setup}.

Figure~\ref{fig:grpo-training} shows three contrasting patterns. 
On the text baseline, GRPO improves accuracy by 3.0 points on 4B (7.8\% $\to$ 10.8\%) and 2.8 points on 8B (10.0\% $\to$ 12.8\%), suggesting that the same training recipe is effective on a well-formed input. 
On the unannotated visual input, the same procedure is essentially ineffective: 4B accuracy moves from 1.4\% to 0.8\% and 8B from 0.2\% to 0.4\%, indicating that the policy cannot discover useful gradient signal from naive image observations. 
On the scaffolded visual input, however, GRPO yields the largest absolute improvements in our study, $+3.8$ points on 4B (1.4\% $\to$ 5.2\%) and $+4.6$ points on 8B (1.2\% $\to$ 5.8\%), both exceeding the corresponding text-baseline gains.
Only the Text Symbols gains have disjoint pre- and post-training intervals (Appendix~\ref{app:bootstrap}); the smaller text-baseline gains are consistent in direction across both sizes.

The pattern suggests visual scaffolding therefore does more than improve zero-shot accuracy.
It exposes task-relevant structure in a form that RL can exploit, avoiding both the compression of text linearization and the perceptual opacity of naive image rendering.

\section{Conclusion}
We investigated how visual presentation shapes VLM performance and failure modes when the underlying reasoning problem is fixed. 
Using SPaRC as a visual spatial planning testbed, we introduced a hierarchy of lightweight visual scaffolds that progressively increase perceptual support while preserving the same board layout, rule configuration, and target solution. 
Across multiple VLMs, these scaffolds substantially improve task-solving accuracy over the original visual board by up to 34.0 percentage points and further make GRPO post-training more effective.

Our analyses further clarify the source of these improvements. 
Object-detection and path-level validity results reveal that scaffolding mainly reduces grounding-related errors while rule satisfaction remains comparatively challenging. 
Thus, visual scaffolding separates grounding-sensitive errors from residual reasoning challenges.

These findings suggest that visual presentation can change both what VLMs can solve and what their failures measure.
Future benchmarks for multimodal planning should control both the underlying reasoning problem and the perceptual interface through which that problem is presented.

Our empirical claims are restricted to SPaRC, and we do not claim that the gains or the scaffolds transfer to natural images, robotics, or document understanding. What is portable is the experimental logic: intervene on the input, hold the task fixed, and observe where the errors move.
The residual errors point to directions for improvement, namely coordinate- or layout-aware visual encoders for the grounding-sensitive part of the gap, and planning-aware post-training or search-based decoding for the rule-satisfaction errors that persist under every scaffold.

\section*{Limitations}

Our scaffolds are hand-designed from SPaRC's known board structure, so each makes explicit a requirement the task already imposes, and the improvements are partly by construction. They are diagnostic interventions rather than a general prompting strategy, and we do not study which scaffold is minimally sufficient or whether they can be built automatically. Because the steps are cumulative rather than orthogonal, we do not read the hierarchy as separating perception from planning, and we count in-image text reading as perception.

All results come from SPaRC, whose grid regularity is both what makes these scaffolds natural and what lets us hold the board layout, rule configuration, and target solution fixed while varying only presentation. We therefore restrict the empirical claims to this benchmark.

Our model pool covers seven open-weight VLMs from 24B to 397B parameters and spans both dense and Mixture-of-Experts architectures, but does not include closed frontier systems such as GPT-5, Gemini 3.1, or Claude, which could in principle respond differently to our scaffolding hierarchy. The trend is consistent across model families (Qwen, Gemma, Llama, GLM, Mistral) and over an order of magnitude in scale, suggesting that the sensitivity to visual presentation reflects current VLM design rather than a quirk of any single family. Using open weights also ensures reproducibility of the results.

Qwen 3.5 397B and GLM 4.6V are run with 4-bit weight quantization to fit our VRAM budget, which likely reduces absolute accuracy for both models. Our central claims are based on within-model comparisons across input representations, and the quantization level is held constant per model. It therefore cannot account for the large gains in scaffolding we observe, including the 34.0 percentage point jump on Qwen 3.5 397B.

The GRPO experiments are limited to Qwen 3 VL 4B Thinking and Qwen 3 VL 8B Thinking since larger reasoning models exceeded our training budget. Consequently, we cannot directly characterize how RL post-training scales for the largest models. However, the pattern we observe is consistent across the two sizes that we trained on. There are near-zero GRPO gains on the Original board compared to up to 4.6 percentage points on Text Symbols. This comparison is the most important one for our claim that perceptual access constrains what RL post-training can teach.

Finally, accuracies are computed over 500 puzzles. The main scaffolding gains are resolvable against the bootstrap intervals in Appendix~\ref{app:bootstrap}, but smaller effects are not, including the GRPO gains on the text baseline (\S\ref{sec:4.6}) and any gain for the two models at the accuracy floor.

\section*{Acknowledgments}

This work was partially supported by the Lower Saxony Ministry of Science and Culture and the VW Foundation. This work used the Scientific Compute Cluster at GWDG, the joint data center of Max Planck Society for the Advancement of Science (MPG) and University of Göttingen. In part funded by the Deutsche Forschungsgemeinschaft (DFG, German Research Foundation) – 405797229.

\bibliography{custom}

\appendix
\crefalias{section}{appendix}%
\crefalias{subsection}{appendix}%
\crefalias{subsubsection}{appendix}%

\section{Model and Hardware Details}
\label{app:models_hardware}

\paragraph{Hardware}
All inference and evaluation procedures, across both the baseline and visual scaffolding settings, were performed on a uniform hardware configuration of 4 NVIDIA A100 GPUs, each equipped with 80GB of VRAM. This provided sufficient memory and compute capacity to run most evaluated VLMs at full precision. For the two largest models, Qwen 3.5 397B and GLM 4.6V, we additionally apply 4-bit weight quantization to fit them within our VRAM budget while preserving their multimodal capabilities. All other models are evaluated without quantization. For GRPO training of the Qwen 3 VL 4B and 8B Thinking model, we used up to four nodes, each with four NVIDIA A100 GPUs.

\paragraph{Evaluated Models}
We evaluate seven vision-language models spanning different sizes, architectures, and training paradigms in order to assess how perceptual accessibility shapes downstream planning behavior across a representative cross-section of current open-weight VLMs. The selection covers both dense and Mixture-of-Experts (MoE) architectures, ranges from 24B-parameter models to a 397B MoE, and includes models from several major model families. The full list is given below:

\begin{itemize}
    \item \raisebox{-0.1\height}{\includegraphics[width=0.9em]{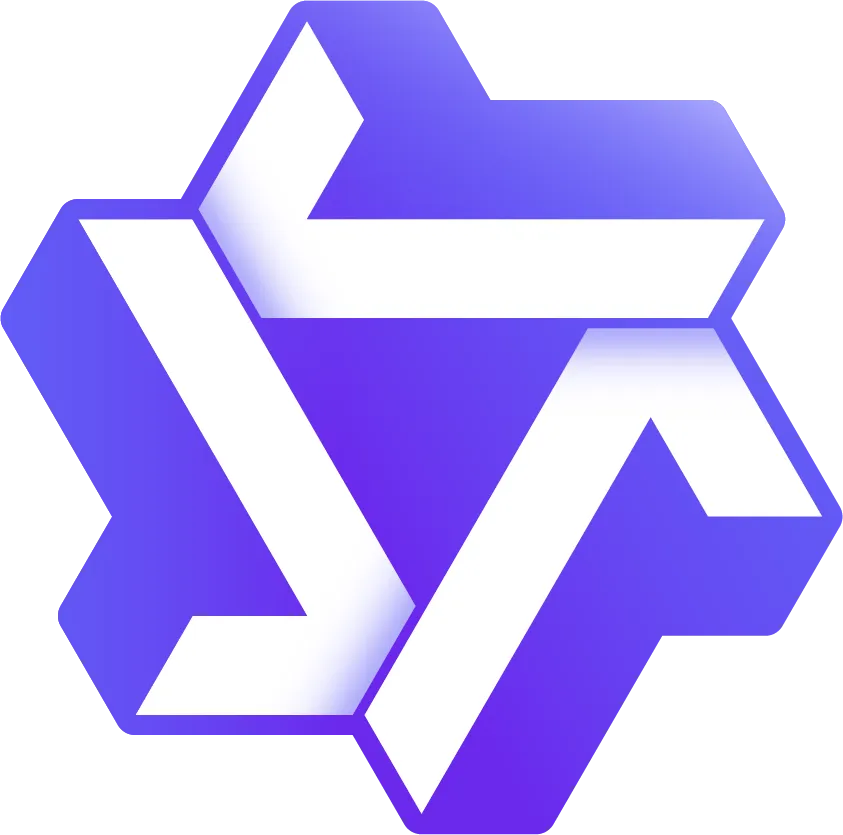}} \textbf{Qwen 3.5 397B (4-bit)}: A large-scale Mixture-of-Experts vision-language model from Alibaba Cloud's Qwen 3.5 generation. We apply 4-bit weight quantization to enable inference within our hardware budget while retaining its multimodal reasoning capability \cite{qwen3.5}.
    \item \raisebox{-0.1\height}{\includegraphics[width=0.9em]{images/logos/qwen.png}} \textbf{Qwen 3.5 27B}: A mid-sized dense vision-language model from the same Qwen 3.5 family, evaluated in full precision and serving as a reference point for the quantized 397B variant within the same model family \cite{qwen3.5}.
    \item \raisebox{-0.1\height}{\includegraphics[width=0.9em]{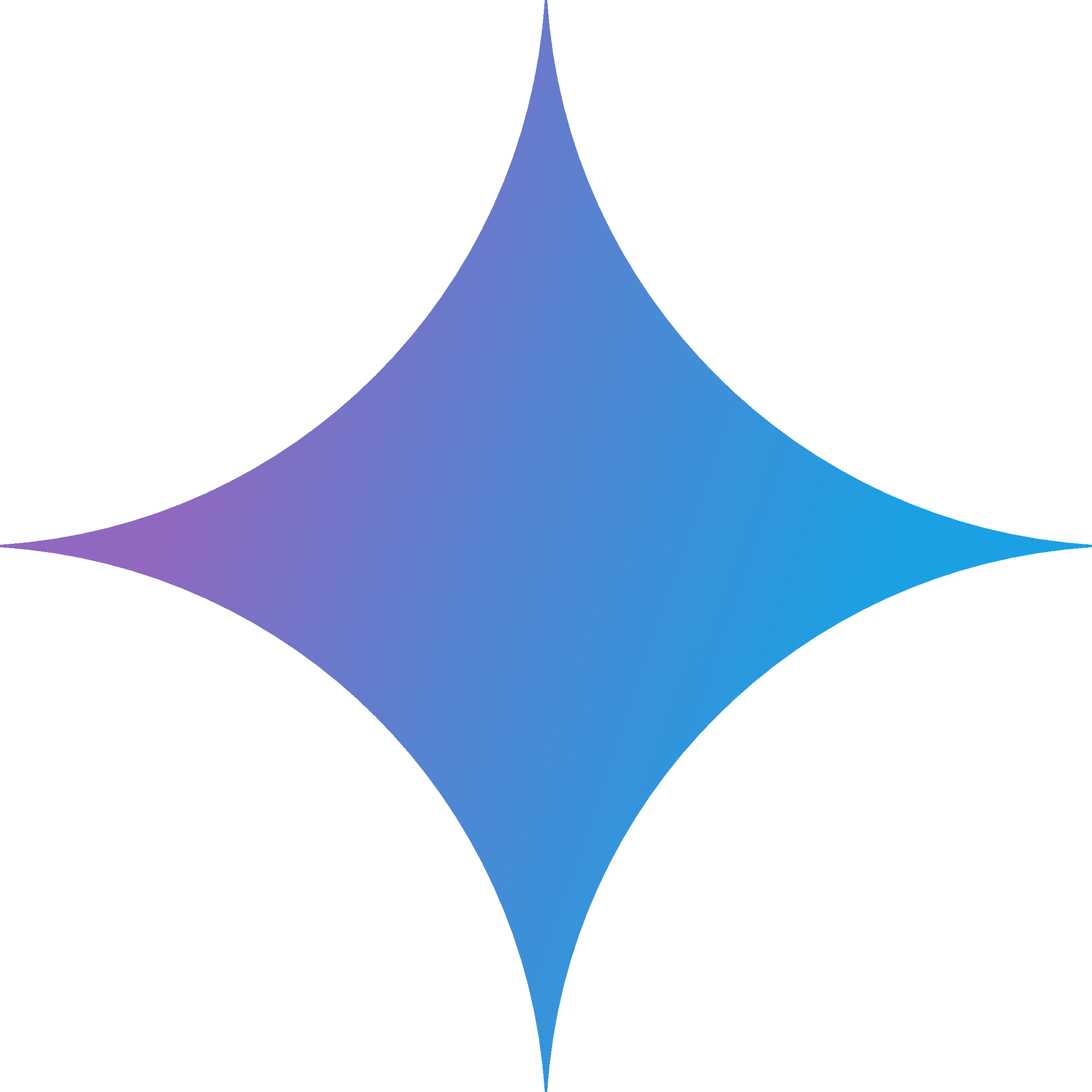}} \textbf{Gemma 4 31B}: Google's latest open-weight multimodal model in the Gemma series, building on Gemini technology with improved visual grounding and long-context reasoning over its predecessor \cite{gemma4}.
    \item \raisebox{-0.1\height}{\includegraphics[width=0.9em]{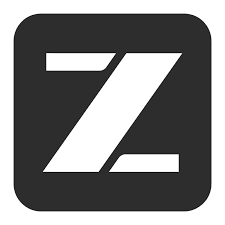}} \textbf{GLM 4.6V (4-bit)}: Zhipu AI's vision-language model from the GLM 4.6 series, designed for multimodal understanding tasks. We apply 4-bit weight quantization to fit it within our VRAM budget \cite{vteam2026glm45vglm41vthinkingversatilemultimodal}.
    \item \raisebox{-0.1\height}{\includegraphics[width=0.9em]{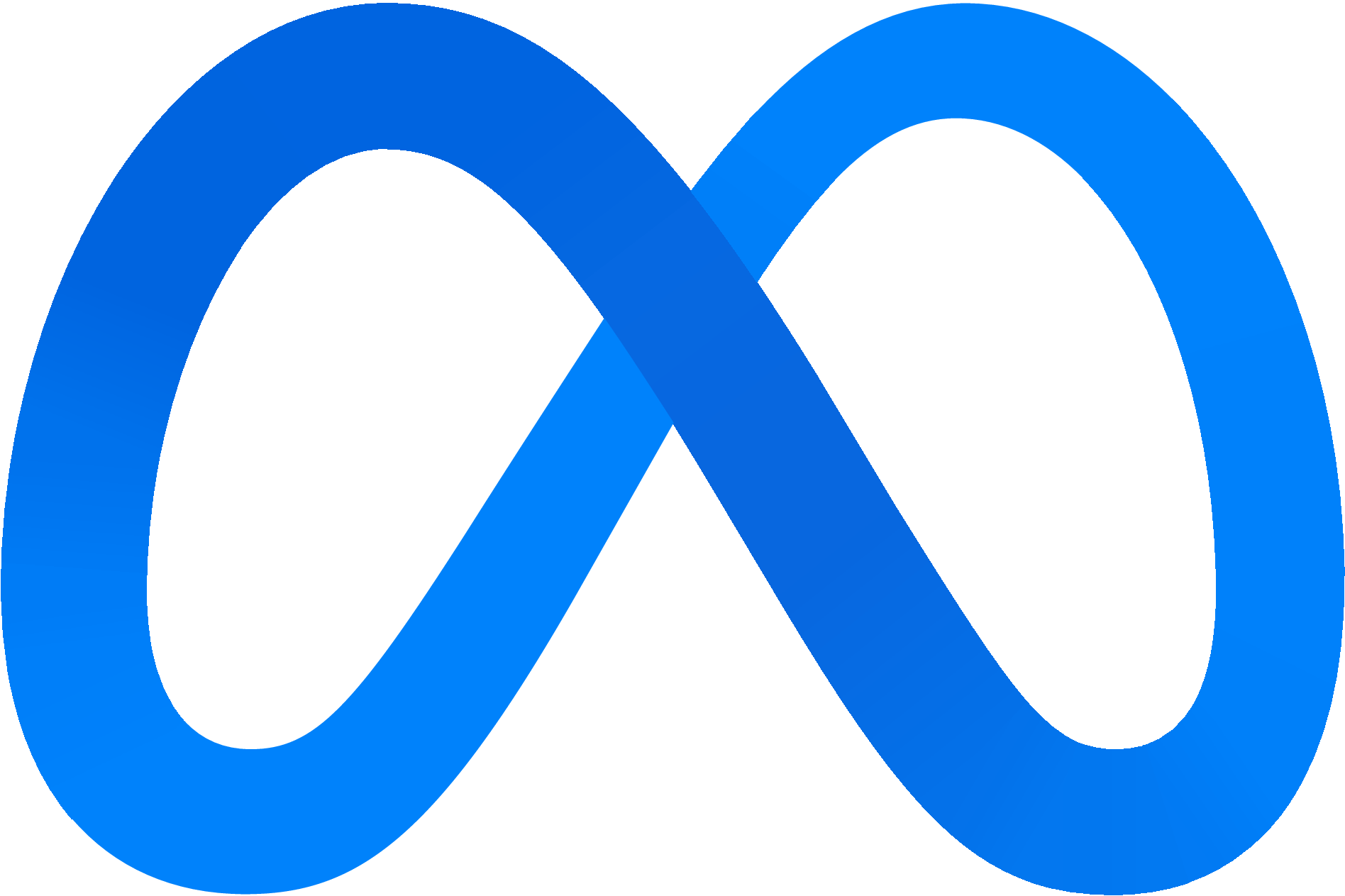}} \textbf{Llama 4 Scout}: Meta's natively multimodal Mixture-of-Experts model from the Llama 4 family, with 17B active parameters and a long context window, designed as the efficient tier of the Llama 4 release \cite{llama4}.
    \item \raisebox{-0.1\height}{\includegraphics[width=0.9em]{images/logos/gemma.png}} \textbf{Gemma 3 27B}: Google's previous-generation open-weight multimodal model, built on Gemini technology and designed for long-context multimodal reasoning. It serves as a within-family comparison point against Gemma 4 31B \cite{gemmateam2025gemma3technicalreport}.
    \item \raisebox{-0.1\height}{\includegraphics[width=0.9em]{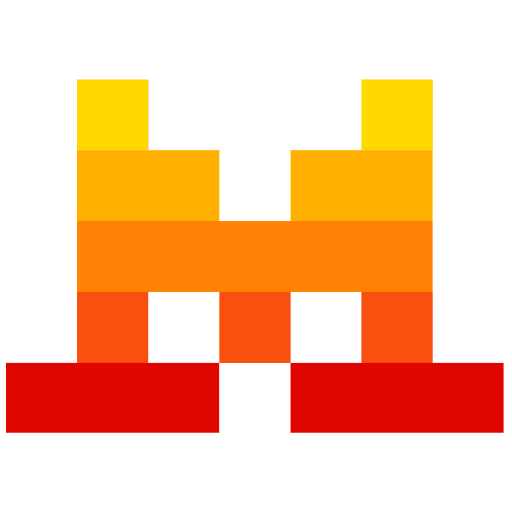}} \textbf{Mistral Small 3.2}: A 24B-parameter multimodal model from Mistral AI in the Mistral Small line, included as a smaller-scale dense vision-language baseline \cite{mistralsmall32}.
\end{itemize}

Together, these seven models allow us to study how the effect of input-side visual scaffolding varies with model scale, architecture, and training paradigm, while keeping the underlying planning task fixed.

\paragraph{Compute Budget}
Table~\ref{tab:compute_budget} reports the aggregate inference cost for the seven evaluated VLMs, summed over all controlled board representations, prompt variants, the visual-degradation runs, and the object-detection evaluations. Token counts are taken from the per-run \texttt{token\_usage} fields emitted by our evaluation pipeline; wall-clock hours are taken from the corresponding \texttt{evaluation\_duration\_seconds} entries. A100-hours are reported as wall-clock hours multiplied by four (the fixed hardware configuration described above).

\begin{table*}[h]
    \centering
    \small

     \begin{tabular}{cl|rrrrr}
        \toprule
        & \textbf{Model} & \textbf{Prompt (M)} & \textbf{Completion (M)} & \textbf{Total (M)} & \textbf{Wall (h)} & \textbf{A100-h} \\
        \midrule
        \raisebox{-0.2\height}{\includegraphics[width=1em]{images/logos/mistral.png}} & Mistral Small 3.2 & 12.1 & 15.2 & 27.3 & 7.9 & 31.7 \\
        \raisebox{-0.2\height}{\includegraphics[width=1em]{images/logos/gemma.png}} & Gemma 3 27B & 10.5 & 4.5 & 15.0 & 2.6 & 10.4 \\
        \raisebox{-0.2\height}{\includegraphics[width=1em]{images/logos/llama.png}} & Llama 4 Scout & 14.6 & 4.6 & 19.2 & 1.1 & 4.2 \\
        \raisebox{-0.2\height}{\includegraphics[width=1em]{images/logos/glm.png}} & GLM 4.6V (4-bit) & 11.7 & 65.1 & 76.8 & 30.9 & 123.8 \\
        \raisebox{-0.2\height}{\includegraphics[width=1em]{images/logos/gemma.png}} & Gemma 4 31B & 21.0 & 38.5 & 59.5 & 16.7 & 66.9 \\
        \raisebox{-0.2\height}{\includegraphics[width=1em]{images/logos/qwen.png}} & Qwen 3.5 27B & 11.2 & 215.4 & 226.6 & 102.6 & 410.3 \\
        \raisebox{-0.2\height}{\includegraphics[width=1em]{images/logos/qwen.png}} & Qwen 3.5 397B (4-bit) & 29.0 & 489.8 & 518.7 & 180.0 & 719.8 \\
        \midrule
        & \textbf{Total} & \textbf{110.0} & \textbf{833.1} & \textbf{943.1} & \textbf{341.8} & \textbf{1{,}367.1} \\
        \bottomrule
    \end{tabular}
        \caption{Per-model token usage and compute consumed on $4\times$A100 (80\,GB). Token counts are in millions; wall-clock and A100-hours are aggregated across all evaluation runs for that model (controlled boards, prompt variants, visual degradations, and object-detection sub-tasks).}
            \label{tab:compute_budget}
\end{table*}

Completion tokens account for roughly $88\%$ of the total, reflecting the long reasoning chains produced by the Qwen 3.5 and GLM 4.6V models on the SPaRC puzzles. Qwen 3.5 397B is the dominant cost contributor at $\approx\!720$ A100-hours, both because it participates in the additional worsening-study runs and because its quantized inference is the slowest per-token. By contrast, Llama 4 Scout terminates quickly with very short outputs and contributes only $\approx\!4$ A100-hours despite being run on the same set of board representations. The reported numbers cover inference only; GRPO training of Qwen 3 VL 4B / 8B Thinking is accounted for separately in Appendix~\ref{app:grpo-setup}.

\section{SPaRC Dataset}
\label{app:sparc_dataset}

SPaRC~\citep{kaesberg-etal-2025-sparc} consists of 1{,}000 2D grid pathfinding puzzles (500 train, 500 test) inspired by the puzzle mechanics of \textit{The Witness}~\citep{witness}.
Each puzzle is an $m \times n$ grid of \textbf{rule cells} with $(x,y)=(0,0)$ at the top-left corner, $x$ increasing rightward and $y$ downward.
Rule cells are surrounded by \textbf{edges} along which the solution path is drawn.
Each puzzle has exactly one \textbf{start point} (large circle) and one \textbf{end point} (edge extension).
The goal is to draw a single, continuous, non-self-intersecting path along edges from start to end that satisfies all rule cell constraints.

\subsection{Rules}
\label{app:sparc_rules}

Seven rule types can appear in SPaRC puzzles:

\long\def\figwindownonum[#1,#2,#3,#4] {%
  \begin{window}[#1,#2,{#3},{\centering#4\par}] }
\def\endfigwindownonum{\end{window}}%

\newcommand{\iconheight}{5ex}

\newcommand{\puzzleitem}[3]{%
  \par\smallskip                    %
  \begin{figwindownonum}[0,l,{
    \includegraphics[width=\iconheight]{#1}},{}]
    \noindent\textbf{#2:} #3%
  \end{figwindownonum}%
  \par\medskip                     %
}

\puzzleitem{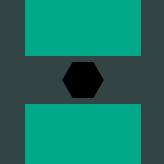}{Item Collection (Dots)}{%
The solution path needs to pass through every dot.
}

\puzzleitem{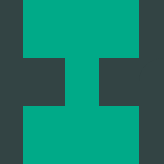}{Path Breaks (Gaps)}{%
The solution path cannot go through any edge segment containing a gap. Gaps act as local barriers.
}

\puzzleitem{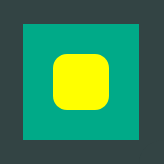}{Color Separation (Stones)}{%
The solution path must be drawn to separate stones of different colors. All stones located within any single enclosed region must be of the same color.
}

\puzzleitem{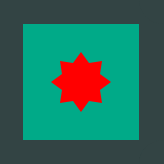}{Pairing (Stars)}{%
Each star must share its region with exactly one other symbol of the same color. No unpaired stars are allowed.
}

\puzzleitem{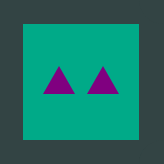}{Edge Count (Triangles)}{%
The solution path must touch the number of edges shown by the triangles in the cell, e.g., two triangles mean the path must touch exactly two edges of that cell.
}

\puzzleitem{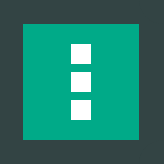}{Shape Fitting (Polyominoes)}{%
If a cell contains a polyomino (poly), the solution path must enclose a region that matches its exact shape and area. The region must not rotate or mirror the poly. Multiple polys can share a region if their shapes fit without overlapping.
}

\puzzleitem{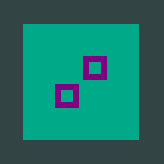}{Shape Subtraction (Ylop)}{%
A ylop must be enclosed in the same region as one or more polys. Its shape and area subtract from the total required by the polys. If a ylop cancels out a poly exactly, that pair imposes no constraint.
}

\subsection{Dataset Creation}
\label{app:dataset_creation}
 
Puzzles are generated by randomly creating an $m \times n$ grid ($m, n \in [2, 6]$), filling approximately half the cells with rules (\textit{rule density}), and placing random start and end points.
A generation-validation loop solves each puzzle via brute-force search over all valid paths.\footnote{Brute-force is necessary because many puzzles fall into NP or NP-Complete complexity classes~\citep{abel2019witnesseswitnessfindingwitnesses}.}
If no solution exists, rule density is decreased; if more than $k{=}50$ solutions exist, density is increased, and the puzzle is regenerated.
 
\paragraph{Difficulty estimation.} Puzzle complexity is quantified by a weighted sum of the number of distinct rule types, total rule cells, rule density, grid size, and estimated rule interactions, normalized to a 1 (easiest) to 5 (hardest) scale.
The test set contains 86 puzzles at level~1, 118 at level~2, 121 at level~3, 86 at level~4, and 89 at level~5.

\section{Comparison with the SPaRC Text Baseline}
\label{app:text-baseline}

\begin{figure}[t]
\centering
\includegraphics[width=\linewidth]{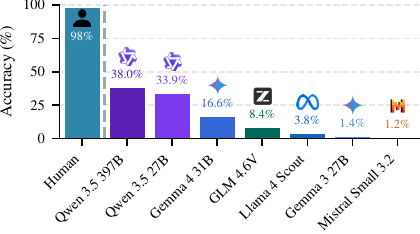}
\caption{Accuracy on the SPaRC text-formulation
baseline for human solvers from \citet{kaesberg-etal-2025-sparc} and the seven evaluated VLMs.}
\label{fig:text-baseline}
\end{figure}

Figure~\ref{fig:text-baseline} reports accuracy
on the original SPaRC text-formulation baseline for human
solvers from \citet{kaesberg-etal-2025-sparc} and the seven evaluated VLMs. Humans reach $98\%$.
The strongest VLM (Qwen~3.5~397B) reaches $38.0\%$, and
four of seven models score below $10\%$. The human-model
gap on a purely textual presentation shows that SPaRC is
non-trivial independent of visual encoding.

Comparing against the best scaffolded visual result per
model from \S\ref{sec:4.2}: Qwen~3.5~397B improves
from $38.0\%$ to $48.4\%$ ($+10.4$\,pp, Text Symbols),
Gemma~4 31B from $16.6\%$ to $26.2\%$ ($+9.6$\,pp, Text
Symbols), Qwen~3.5 27B from $33.9\%$ to $40.0\%$
($+6.1$\,pp, Cell Coordinates), and GLM~4.6V from $8.4\%$
to $12.8\%$ ($+4.4$\,pp, Cell Coordinates). The three
weakest models (Llama~4 Scout, Gemma~3 27B, Mistral
Small~3.2) stay within $\pm 1$\,pp of their text baseline,
all near the floor. The visual-over-text gain scales with
model capability, matching the scaffolding-gain pattern in
\S\ref{sec:4.2}.

\section{Recovery Variants}
\label{app:recovery_variants}

\begin{figure*}[htbp]
    \centering
    \includegraphics[width=0.85\textwidth]{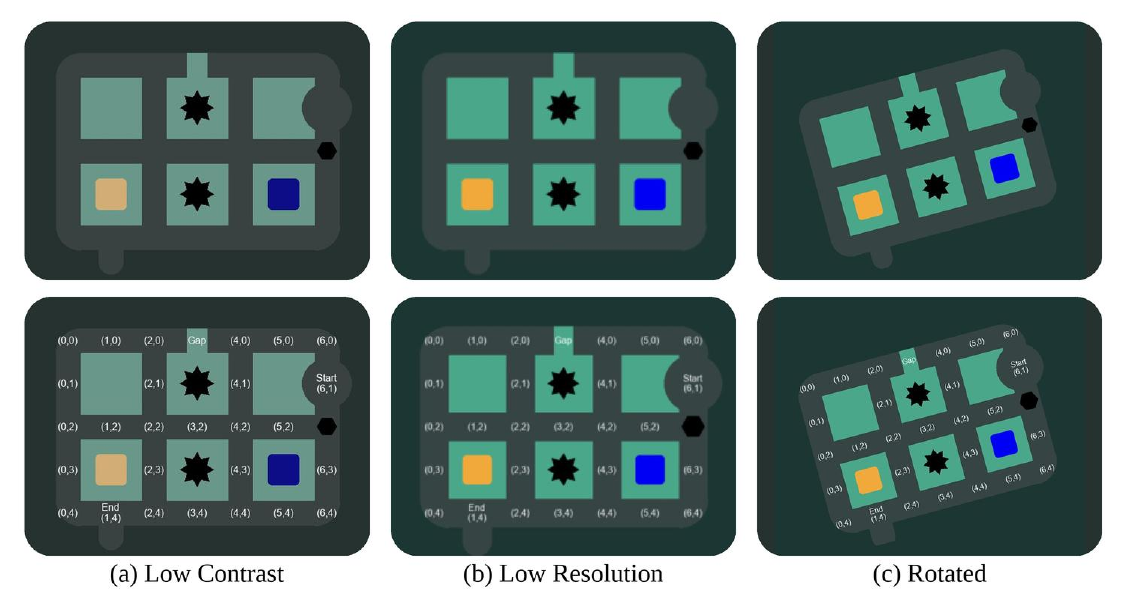}
    \caption{The three visual degradations (low contrast, low resolution, and rotation) shown above their recovery counterparts, in which Cell Coordinates are overlaid on the degraded board.}

    \label{fig:worsenings_recovery}
\end{figure*}

Figure~\ref{fig:worsenings_recovery} shows the three visual degradation variants introduced in Section~\ref{sec:3.3} alongside their recovery counterparts. Each recovery variant overlays the Cell Coordinates representation on the degraded board, leaving the planning task and degradation parameters unchanged. This isolates the contribution of explicit spatial indexing from any change to the visual signal itself.

\section{Additional Experiments}
\label{app:additional}

\subsection{Bootstrap Confidence Intervals}
\label{app:bootstrap}

Accuracies are computed over 500 test puzzles, so we report $95\%$ bootstrap percentile intervals obtained by resampling puzzle IDs with replacement ($200{,}000$ replicates, seed 20260821).
Table~\ref{tab:bootstrap} covers every model and condition and Table~\ref{tab:bootstrap_grpo} the GRPO runs. Table~\ref{tab:bootstrap_paired} reports differences to each model's best scaffold, where each replicate resamples puzzle IDs once and evaluates both conditions on that same resample.
The gains are resolvable for the five models above the accuracy floor. Gemma~3~27B and Mistral~Small~3.2 stay below $1\%$ on every visual condition and show no resolvable gain from any scaffold.

\begin{table}[htbp]
\centering
\small
\setlength{\tabcolsep}{4pt}
\begin{tabular}{llrc}
\toprule
\textbf{Model} & \textbf{Setting} & \textbf{Acc.} & \textbf{95\% CI} \\
\midrule
Gemma 3 27B & Original & \phantom{0}0.0 & [\phantom{0}0.0, \phantom{0}0.0] \\
 & S/E & \phantom{0}0.0 & [\phantom{0}0.0, \phantom{0}0.0] \\
 & Axis & \phantom{0}0.0 & [\phantom{0}0.0, \phantom{0}0.0] \\
 & Axis+S/E & \phantom{0}0.0 & [\phantom{0}0.0, \phantom{0}0.0] \\
 & Cell Coord. & \phantom{0}0.0 & [\phantom{0}0.0, \phantom{0}0.0] \\
 & Text Sym. & \phantom{0}0.6 & [\phantom{0}0.0, \phantom{0}1.4] \\
 & Text base. & \phantom{0}1.4 & [\phantom{0}0.4, \phantom{0}2.6] \\
\midrule
Gemma 4 31B & Original & \phantom{0}2.2 & [\phantom{0}1.0, \phantom{0}3.6] \\
 & S/E & \phantom{0}9.2 & [\phantom{0}6.8, 11.8] \\
 & Axis & \phantom{0}9.8 & [\phantom{0}7.2, 12.4] \\
 & Axis+S/E & 22.4 & [18.8, 26.2] \\
 & Cell Coord. & 25.0 & [21.2, 28.8] \\
 & Text Sym. & 26.2 & [22.4, 30.2] \\
 & Text base. & 16.6 & [13.4, 20.0] \\
\midrule
Qwen 3.5 27B & Original & \phantom{0}8.2 & [\phantom{0}5.8, 10.6] \\
 & S/E & 23.0 & [19.4, 26.8] \\
 & Axis & 12.0 & [\phantom{0}9.2, 15.0] \\
 & Axis+S/E & 35.0 & [30.8, 39.2] \\
 & Cell Coord. & 40.0 & [35.8, 44.4] \\
 & Text Sym. & 37.6 & [33.4, 41.8] \\
 & Text base. & 33.9 & [29.7, 38.2] \\
\midrule
Qwen 3.5 397B & Original & 14.4 & [11.4, 17.6] \\
 & S/E & 32.6 & [28.6, 36.8] \\
 & Axis & 16.8 & [13.6, 20.2] \\
 & Axis+S/E & 45.8 & [41.4, 50.2] \\
 & Cell Coord. & 47.8 & [43.4, 52.2] \\
 & Text Sym. & 48.4 & [44.0, 52.8] \\
 & Text base. & 38.0 & [33.8, 42.2] \\
\midrule
Llama 4 Scout & Original & \phantom{0}0.0 & [\phantom{0}0.0, \phantom{0}0.0] \\
 & S/E & \phantom{0}0.4 & [\phantom{0}0.0, \phantom{0}1.0] \\
 & Axis & \phantom{0}0.2 & [\phantom{0}0.0, \phantom{0}0.6] \\
 & Axis+S/E & \phantom{0}0.6 & [\phantom{0}0.0, \phantom{0}1.4] \\
 & Cell Coord. & \phantom{0}3.2 & [\phantom{0}1.8, \phantom{0}4.8] \\
 & Text Sym. & \phantom{0}2.0 & [\phantom{0}0.8, \phantom{0}3.4] \\
 & Text base. & \phantom{0}3.8 & [\phantom{0}2.2, \phantom{0}5.6] \\
\midrule
Mistral Small 3.2 & Original & \phantom{0}0.0 & [\phantom{0}0.0, \phantom{0}0.0] \\
 & S/E & \phantom{0}0.4 & [\phantom{0}0.0, \phantom{0}1.0] \\
 & Axis & \phantom{0}0.0 & [\phantom{0}0.0, \phantom{0}0.0] \\
 & Axis+S/E & \phantom{0}0.4 & [\phantom{0}0.0, \phantom{0}1.0] \\
 & Cell Coord. & \phantom{0}0.2 & [\phantom{0}0.0, \phantom{0}0.6] \\
 & Text Sym. & \phantom{0}0.4 & [\phantom{0}0.0, \phantom{0}1.0] \\
 & Text base. & \phantom{0}1.2 & [\phantom{0}0.4, \phantom{0}2.2] \\
\midrule
GLM 4.6V & Original & \phantom{0}1.0 & [\phantom{0}0.2, \phantom{0}2.0] \\
 & S/E & \phantom{0}2.6 & [\phantom{0}1.4, \phantom{0}4.0] \\
 & Axis & \phantom{0}1.2 & [\phantom{0}0.4, \phantom{0}2.2] \\
 & Axis+S/E & \phantom{0}7.6 & [\phantom{0}5.4, 10.0] \\
 & Cell Coord. & 12.8 & [10.0, 15.8] \\
 & Text Sym. & 10.6 & [\phantom{0}8.0, 13.4] \\
 & Text base. & \phantom{0}8.4 & [\phantom{0}6.0, 10.8] \\
\bottomrule
\end{tabular}
\caption{Task accuracy (\%) with $95\%$ bootstrap confidence intervals over the test split ($B = 200{,}000$). \textit{Text base.} is the non-visual text formulation; \textit{Text Sym.} renders rule symbols as words.}
\label{tab:bootstrap}
\end{table}

\begin{table}[htbp]
\centering
\small
\setlength{\tabcolsep}{3pt}
\begin{tabular}{lcc}
\toprule
\textbf{Model} & \textbf{from Original} & \textbf{from Text base.} \\
\midrule
Mistral 3.2  & \phantom{0}$+0.4$ [0.0, 1.0] & $-0.8$ [$-1.8$, 0.0] \\
Gemma 3 27B  & \phantom{0}$+0.6$ [0.0, 1.4] & $-0.8$ [$-2.0$, 0.2] \\
Llama 4      & \phantom{0}$+3.2$ [1.8, 4.8] & $-0.6$ [$-2.2$, 1.0] \\
GLM 4.6V     & $+11.8$ [9.0, 14.8] & $+4.4$ [1.4, \phantom{0}7.4] \\
Gemma 4 31B  & $+24.0$ [20.2, 27.8] & $+9.6$ [5.8, 13.4] \\
Qwen 27B     & $+31.8$ [27.6, 36.0] & $+6.2$ [2.2, 10.2] \\
Qwen 397B    & $+34.0$ [29.4, 38.6] & $+10.4$ [6.2, 14.6] \\
\bottomrule
\end{tabular}
\caption{Paired differences (percentage points) from the Original board and from the text baseline to each model's best scaffold.}
\label{tab:bootstrap_paired}
\end{table}

\begin{table}[htbp]
\centering
\small
\setlength{\tabcolsep}{3pt}
\begin{tabular}{lcc}
\toprule
\textbf{Setting} & \textbf{Before} & \textbf{After} \\
\midrule
\multicolumn{3}{l}{\textit{Qwen 3 VL 4B}} \\
Text base. & \phantom{0}7.8 [5.6,10.2] & 10.8 [\phantom{0}8.2,13.6] \\
Original   & \phantom{0}1.4 [0.4,\phantom{0}2.6] & \phantom{0}0.8 [\phantom{0}0.2,\phantom{0}1.6] \\
Text Sym.  & \phantom{0}1.4 [0.4,\phantom{0}2.6] & \phantom{0}5.2 [\phantom{0}3.4,\phantom{0}7.2] \\
\midrule
\multicolumn{3}{l}{\textit{Qwen 3 VL 8B}} \\
Text base. & 10.0 [7.4,12.8] & 12.8 [10.0,15.8] \\
Original   & \phantom{0}0.2 [0.0,\phantom{0}0.6] & \phantom{0}0.4 [\phantom{0}0.0,\phantom{0}1.0] \\
Text Sym.  & \phantom{0}1.2 [0.4,\phantom{0}2.2] & \phantom{0}5.8 [\phantom{0}3.8,\phantom{0}8.0] \\
\bottomrule
\end{tabular}
\caption{Task accuracy (\%) before and after GRPO post-training, with $95\%$ bootstrap confidence intervals.}
\label{tab:bootstrap_grpo}
\end{table}

\subsection{Grid Lines versus Coordinate Labels}
\label{app:grid_label_ablation}

The Axis Labels scaffold changes two things at once. It draws grid lines over the board, and it prints coordinate labels in a margin around it.
To separate visual structure from the labels themselves, we decompose it into three variants on the same 500 test puzzles: grid lines only, drawn with equal margins on all four sides; grid lines with the label margin reserved but left blank; and the full Axis Labels variant.

Table~\ref{tab:grid_label_ablation} reports task-solving accuracy for the three strongest models.
Grid lines alone give small gains over the Original board ($+1.4$, $+0.2$, and $+2.6$ pp). Reserving the margin but leaving it blank hurts every model, falling below the Original board in all three cases. Printing the labels into the same margin recovers and exceeds it ($+2.4$ to $+7.6$ pp).

An error analysis explains the blank-margin result.
We measure the rate of $\pm1$ coordinate offsets in the predicted start and end nodes, that is, predictions naming a cell one step away from the correct one along either axis.
This rate is $4$--$5\%$ with explicit labels and rises to $17$--$46\%$ with the blank margin. Removing the margin and keeping only the grid lines brings it back to $6$--$14\%$ for the two Qwen models, while Gemma~4~31B stays shift-prone at $35\%$.

An uninformative reserved margin therefore disrupts cell indexing, since the model appears to treat the blank band as part of the board and offsets its indices accordingly.
Explicit labels instead act as coordinate anchors, and they matter most for the weaker models. The gains come from explicit spatial grounding cues rather than from generic visual modification.

\begin{table}[h]
    \centering
    \small
    \begin{tabular}{lrrrr}
        \toprule
        \textbf{Model} & \textbf{Orig.} & \textbf{+grid} & \textbf{+gutter} & \textbf{+labels} \\
        \midrule
        Qwen 3.5 397B & 14.4 & 15.8 & 4.4 & \textbf{16.8} \\
        Qwen 3.5 27B  &  8.2 &  8.4 & 2.2 & \textbf{12.0} \\
        Gemma 4 31B   &  2.2 &  4.8 & 1.6 & \phantom{0}\textbf{9.8} \\
        \bottomrule
    \end{tabular}
    \caption{Decomposition of the Axis Labels scaffold. Task-solving accuracy (\%) on the 500-puzzle test split for the Original board, grid lines only (equal margins, no labels), grid lines with the label gutter reserved but left empty, and the full Axis Labels variant.}
    \label{tab:grid_label_ablation}
\end{table}

\subsection{GRPO Post-Training}
\label{app:grpo-setup}
In addition to the input-side interventions in (\S \ref{sec:3.2}) and (\S \ref{sec:3.3}), we study a weight-side intervention. We apply Group Relative Policy Optimization (GRPO) to Qwen 3 VL 4B Thinking and Qwen 3 VL 8B Thinking, two open-weight reasoning models that fit our hardware budget for training. Training uses the SPaRC train split (500 examples), and evaluation uses the held-out test split. The reward is based on whether the predicted path satisfies all validity conditions defined in (\S \ref{sec:3.4}). Hyperparameters and compute details are provided in \Cref{app:models_hardware}.
To isolate the role of perceptual access in trainability, we evaluate GRPO under three input conditions, namely the SPaRC text baseline, the Original visual board, and the Text Symbols scaffold. For each condition, the model is trained and evaluated on the same input representation, so any difference in GRPO gain across conditions reflects how the input format affects the training signal. We use this comparison as a diagnostic for how perceptual access interacts with RL post-training.

GRPO training of the Qwen 3 VL 4B and 8B Thinking models was run on up to four $4\times $A100 nodes (16 GPUs in total) and required approximately 30\,h and 60\,h of wall-clock time, respectively. This corresponds to on the order of 500 and 1{,}000 A100-hours, bringing the combined GRPO training budget to roughly 1{,}500 A100-hours on top of the inference cost reported in Table~\ref{tab:compute_budget}.

\subsection{Averaged Confusion Matrix}
\label{app:avg_confusion}

\begin{figure}[h]
\centering
\includegraphics[width=\columnwidth]{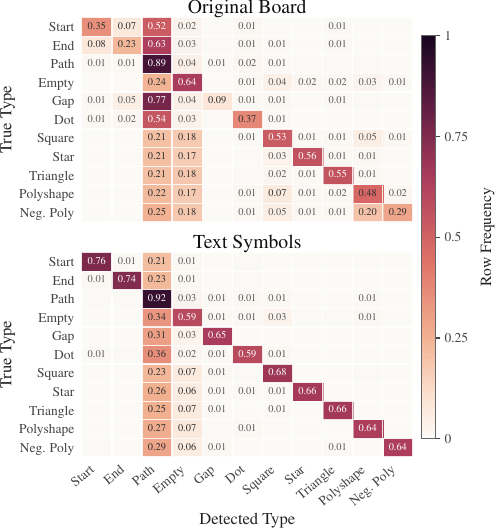}
\caption{Per-type detection accuracy and confusion patterns averaged across all seven evaluated models, for the Original board (top) and Text Symbols representation (bottom).}
\label{fig:avg_confusion}
\end{figure}

Figure~\ref{fig:od_confusion_comparison} in §\ref{sec:4.3} shows per-type confusion patterns for the strongest model, Qwen~3.5~397B.
To check that those patterns are not specific to a single model, Figure~\ref{fig:avg_confusion} reports the same analysis averaged across all seven evaluated VLMs. On the Original board, all symbol types are heavily confused with the dominant Path class: diagonal correctness drops as low as $0.23$ for End, $0.29$ for Negative Polyshape, and $0.37$ for Dot, with most of the mass on the Path column. 
The Text Symbols representation produces a substantially sharper diagonal across models. A residual fraction of cells is still labeled Path or Empty ($\sim$0.20--0.34 for several rule types), which is consistent with weaker models defaulting to background classes even when symbols are made explicit; the dominant gain, however, comes from disambiguating the symbol classes themselves.

\subsection{Correlation with Average Rule Detection Accuracy}
\label{app:avg_correlation}

\begin{figure}[h]
\centering
\includegraphics[width=\columnwidth]{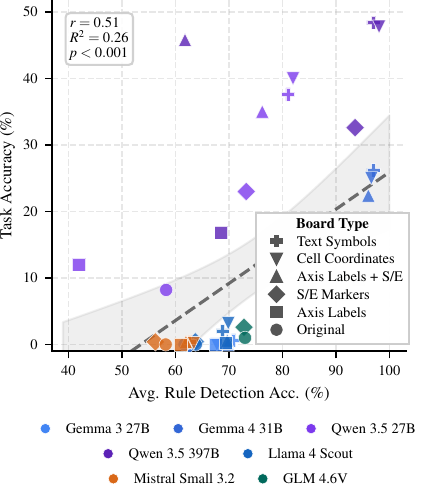}
\caption{Correlation between average rule detection accuracy and end-to-end task-solving accuracy across models and board representations.}
\label{fig:avg_correlation}
\end{figure}

In §\ref{sec:4.3} (Figure~\ref{fig:od_task_correlation}) we report the correlation between Exact-Match Board Accuracy and end-to-end task accuracy. As a complement, Figure~\ref{fig:avg_correlation} shows the same correlation using Average Rule Detection Accuracy as the perception metric. The relationship remains positive and statistically significant ($r=0.51$, $R^2=0.26$, $p<0.001$), but is notably weaker than the exact-match correlation ($r=0.96$). This contrast is informative: solving a puzzle requires all task-relevant board elements to be perceived correctly, so a strict, board-level perception measure aligns more tightly with downstream success than per-rule averages, which are dominated by the many Empty and Path cells and can stay high even when a few critical symbols are mislabelled.

\subsection{Token Usage}
\label{app:token_usage}

\begin{figure}[h]
\centering
\includegraphics[width=\columnwidth]{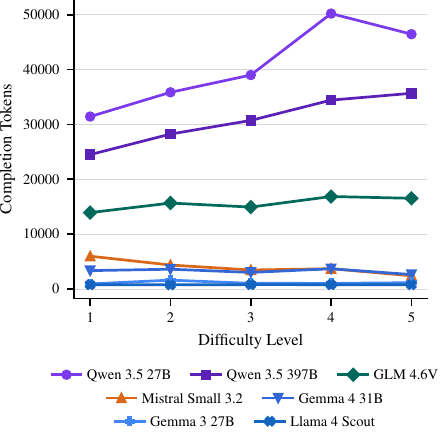}
\caption{Completion tokens per puzzle, stratified by difficulty
level, for each evaluated model on the Original visual input.}
\label{fig:token_usage}
\end{figure}

Figure~\ref{fig:token_usage} reports the average number of completion tokens generated per puzzle, stratified by difficulty level. Inference budgets vary by more than an order of magnitude across model families.
The two Qwen~3.5 models scale their completion length with difficulty, rising from $\sim$31k to $\sim$50k tokens for Qwen~3.5~27B and from $\sim$24k to $\sim$36k for Qwen~3.5~397B. GLM~4.6V uses a moderate and roughly constant budget ($\sim$14k--17k tokens), while Mistral~Small~3.2, Gemma~3~27B, Gemma~4~31B, and Llama~4~Scout stay below $\sim$6k tokens at all difficulties and do not visibly scale with difficulty.

Two observations connect this picture to the scaffolding results in §\ref{sec:4.2}. First, the two models that already invest the largest compute budgets are also the ones with the largest scaffolding gains ($+31.8$ pp for Qwen~3.5~27B and $+34.0$ pp for Qwen~3.5~397B). Even extensive chain-of-thought reasoning does not let these models overcome the grounding difficulty of the Original board; once spatial structure is made explicit, the same reasoning capacity translates into substantially better task accuracy. Second, Gemma~4~31B achieves a comparable $+24.0$ pp gain with completion budgets below $\sim$4k tokens. This shows that the benefit of scaffolding does not require extended reasoning. 
Models with short responses can still benefit from improved visual representations to produce correct solutions. Together, these patterns indicate that perceptual access acts upstream of inference-time compute: improving it benefits both long-reasoning and short-reasoning models, and it is not a substitute for reasoning capacity.

\subsection{Object Detection under Visual Degradation}
\label{app:od_degradation}
\begin{figure}[h]
\centering
\includegraphics[width=\columnwidth]{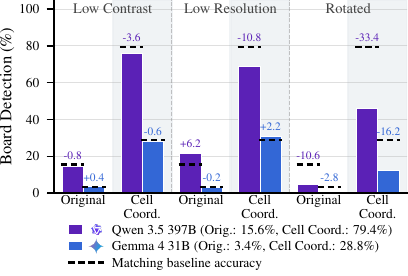}
\caption{Object-detection accuracy under visual degradation and scaffolded recovery settings.}
\label{fig:od_degradation}
\end{figure}

Figure~\ref{fig:od_degradation} reports the change in Full Board Rule Detection Accuracy under the three visual degradation settings introduced in §\ref{sec:4.5}, both on the Original board and after overlaying Cell Coordinates, for the two strongest models (Qwen~3.5~397B and Gemma~4~31B).
Two patterns stand out. First, low contrast and low resolution have only a marginal effect on rule detection on the Original board (all $|\Delta|\!\leq\!6.2$ percentage points), in line with the similarly small changes in end-to-end accuracy under the same degradations in Figure~\ref{fig:worsening_comparison}. Symbol perception is therefore largely robust to the contrast and resolution reductions considered here. Second, rotation reduces rule detection substantially, and the Cell Coordinates overlay degrades sharply on rotated boards ($-33.4$ pp for Qwen~3.5~397B, $-16.2$ pp for Gemma~4~31B); a similar but smaller drop appears under low resolution with Cell Coordinates for Qwen~3.5~397B ($-10.8$ pp). A degraded or rotated coordinate overlay is itself misaligned with the model's canonical reading order and introduces additional visual clutter that competes with, rather than supports, rule recognition. These rule-detection drops on the Cell Coord overlay markedly exceed the corresponding end-to-end accuracy drops (e.g.\ $-9.0$ pp end-to-end vs.\ $-33.4$ pp rule detection for Qwen~3.5~397B under rotation), indicating that the scaffolding's contribution to rule-level perception is the component most affected by geometric transformations, even where downstream reasoning partially compensates. Together, these observations refine the interpretation from §\ref{sec:4.4}. Cell-level scaffolding primarily supports spatial grounding, and the scaffold itself becomes a liability when the board is rotated.

\section{Prompts}

\input{tex/prompts_appendix}
\onecolumn
\section{AI-Usage Card}
\makeAIUsageCard

\input{annotation.tex}

\end{document}

%% file: tex/prompts_appendix.tex
This section lists the three prompts used in the image-grounded evaluation.
All three are presented to the model together with a rendered image of the puzzle.
The \emph{default} prompt (Appendix~\ref{app:prompt-default}) is a compact baseline that lists the rules and the expected output format.
The \emph{improved} prompt (Appendix~\ref{app:prompt-improved}) extends the default with explicit definitions of path cells, rule cells, and regions, and adds worked examples to clarify region detection.
The \emph{object-detection} prompt (Appendix~\ref{app:prompt-detection}) replaces the solving task with a perception-only task: the model must reconstruct the symbolic grid representation of the puzzle from the rendered image, which lets us isolate visual grounding errors from path-finding errors.

The object-detection prompt embeds the full symbol legend and color codes; the default and improved prompts rely on the rendered image plus the descriptions given in their listings.
Curly-brace tokens (e.g.\ \texttt{\{cols\}}, \texttt{\{rows\}}) are placeholders filled in at runtime: \texttt{\{cols\}} and \texttt{\{rows\}} expand to $2W{+}1$ and $2H{+}1$ for a puzzle with a $W\times H$ grid of rule cells, while \texttt{\{max\_x\}} and \texttt{\{max\_y\}} expand to $2W$ and $2H$.

\subsection{Default Prompt}
\label{app:prompt-default}

The default prompt is the baseline image-grounded prompt.
It describes the goal, the coordinate system, and the puzzle rules in a compact form, and instructs the model to output the solution as a list of \texttt{(x,y)} coordinates after a \texttt{\#\#\#\#} marker.
The block \texttt{\{text\_visualization\}} is replaced by a textual rendering of the puzzle when the textual-representation flag is enabled, and is empty otherwise.

\begin{lstlisting}[basicstyle=\ttfamily\scriptsize, breaklines=true, breakatwhitespace=true, frame=single, columns=fullflexible, keepspaces=true, xleftmargin=0.5em, xrightmargin=0.5em]
You are an expert spatial reasoning AI specializing in solving puzzles from the game 'The Witness'.
Your task is to solve the puzzle in the image by finding a valid line from the Start Node to the End Node.

The image shows a Witness puzzle grid of size {cols}x{rows}. In this puzzle:
- The solution is a continuous line from the start circle to the end marker
- The line travels along grid edges, connecting adjacent nodes horizontally or vertically
- The line cannot visit the same node twice
- The line must satisfy all constraints represented by the symbols on the grid
- The line can not be placed on rule cells
- The line can only travel 1 cell per step (no diagonal moves and provide each step as a separate coordinate)

COORDINATE SYSTEM:
- Nodes are indexed (x, y) where (0,0) is the top-left node
- x increases to the right, y increases downward
- The grid cells have rule symbols located at cells with all odd coordinates
- The line goes AROUND cells containing rules, forming boundaries
- Both line and rule cells are on the same grid. Therefore each intersection has a distance of 2 to the next intersection.

SOLVING RULES:
1. Draw a continuous line from the START NODE (big circle on the line) to the END NODE (rounded end) without visiting the same node twice.
2. The line can only be placed on valid path cells.
3. The line acts as a boundary, potentially dividing the grid cells into one or more distinct regions.
4. All rules associated with symbols on the grid must be satisfied:
   - Dots: The line MUST pass through each dot.
   - Colored squares: All squares within a single region created by the line must be the same color. Different colored squares MUST be separated into different regions by the line.
   - Colored stars: Each star must be paired with EXACTLY one other element of the same color in a region. Other colors are ignored.
   - Triangles: The line must touch EXACTLY the number of edges specified by the number of triangles in that cell (edges are top, right, bottom, left of the cell).
   - Tetris-like polyomino shapes: The region containing this symbol must be shaped EXACTLY like the defined polyshape.
   - Negative polyshapes: These cancel out regular polyshapes if they overlap.

Text description of the puzzle:
{text_visualization}

Analyze the puzzle image carefully and determine the solution path.
First, explain your reasoning step-by-step, including key deductions and constraint checks made along the way.
Then, provide the final solution as a sequence of node coordinates in (x, y) format, starting with the start node and ending with the end node, after this string: "####". DON'T SKIP ANY intermediate nodes (the distance between each node must be 1).
Example coordinate list: [(0,0), (1,0), (2,0), (2,1), ...]
\end{lstlisting}

\subsection{Improved Prompt}
\label{app:prompt-improved}

The improved prompt expands the default with explicit definitions of path cells, rule cells, and regions; it spells out the parity convention for rule-cell coordinates and includes a worked example for a non-trivial region.
It also separates positive and negative polyshapes and gives the precise output format.

\paragraph{Board variants.}
\phantom{ }
The placeholders \texttt{\{start\_marker\}} and \texttt{\{end\_marker\}} are replaced by the strings `` with an `S' on it'' and ``, indicated by the letter `E''' respectively when the board configuration includes start/end markers (i.e.\ \texttt{start\_end\_marked} appears in the \texttt{board\_type}), and are empty otherwise.

\begin{lstlisting}[basicstyle=\ttfamily\scriptsize, breaklines=true, breakatwhitespace=true, frame=single, columns=fullflexible, keepspaces=true, xleftmargin=0.5em, xrightmargin=0.5em]
## Objective
You are a specialized AI proficient in spatial reasoning and solving puzzles from the game 'The Witness'. Your goal is to find a valid path (a continuous line) from the specified Start Node to the End Node on the provided image, adhering to all puzzle rules.

## Core Concepts & Grid Basics
* **Grid Dimensions:** The puzzle grid has {cols} columns and {rows} rows.
* **Coordinate System:** Nodes are identified by `(x, y)` coordinates. `(0,0)` is the top-left node. `({max_x},{max_y})` is the bottom-right node. `x` increases to the right, `y` increases downwards. Nodes can either be path cells or rule cells.
* **Path Cells:** Path cells are dark grey and are at all positions where at least one of `x` or `y` is even.
* **Path:** The solution is a single, continuous line connecting adjacent path cells either horizontally or vertically. Each step has a distance 1.
* **No Revisits:** The path **CANNOT** visit the same node more than once.
* **Valid Path Cells:** The path travels along the grid lines (edges between nodes). It can only occupy positions on the path (these correspond to positions with at least one even coordinate and no gap).
* **Rule Cells:** Light green cells are rule cells and have coordinates where both `x` and `y` are odd. The path goes *around* these rule cells, never *on* them. Rule cells can contain rule symbols (square, star, triangles, polyshapes) but can also be empty (no constraint).
* **Regions:** The drawn path (in combination with the grid border) divides the rule cells into one or more distinct enclosed areas (regions). Many rules apply based on the contents of these regions.

## Detailed Solving Rules
The drawn path must satisfy **ALL** applicable constraints:

1. **Path Constraints:**
* Path **MUST** start at `Start Node` (big circle on the line in the same color as the grey path{start_marker}) and end at `End Node` (node from which you can escape the grid to the rounded path outside{end_marker}). Both these nodes are guaranteed to be on valid path cells on the edge of the board (at least one of `x` and `y` has to be an even number).
* Path connects adjacent nodes (horizontal/vertical moves only).
* Nodes **CANNOT** be revisited.
* Path **MUST** pass through all Dots (black hexagons) that lay on the path cells.
* Path **CANNOT** pass through any Gap in the path cells (a Gap is the absence of a path at a cell which would normally be a path cell).

2. **Region-Based Rules** (Apply to areas enclosed by the path):
* **Squares:** All squares within a single region **MUST** be the same color. Squares of different colors **MUST** be separated into different regions by the path.
* **Stars:** Within a single region, each star symbol **MUST** be paired with exactly **ONE** other element (star or square) *of the same color*. Other colors within the region are irrelevant to this specific star's rule.
* **Polyshapes** (one or multiple filled squares in a specific arrangement): The region containing this symbol **MUST** be in the specified shape (defined by the specific Polyshape arrangement). The shape must fit entirely within the region's boundaries. If multiple positive polyshapes are in one region, the region must accommodate their combined, non-overlapping forms.
* **Negative Polyshapes** (one or multiple unfilled squares in a specific arrangement): These "subtract" shape requirements, typically within the same region as corresponding positive polyshapes. A negative polyshape cancels out a positive polyshape of the exact same shape and color within that region. If all positive shapes are canceled, the region has no shape constraint. A negative shape is only considered 'used' if it cancels a positive one. Negative shapes can sometimes rationalize apparent overlaps or boundary violations of positive shapes if interpreted as cancellations.
* **Simple Example Region:** If the a path goes through (2,0), (2,1), (2,2), (1,2), (0,2), then the rule cell (1,1) is enclosed into a region containing only itself since the grid border does the rest of the enclosure. To double check: There exist no nonvisited path cells (legal or non legal) that connect (1,1) to any other rule cell.
* **Complex Example Region:** Suppose the grid is 7x7 with rule cells at (1,1), (1,3), (1,5), (3,1), (3,3), (3,5), (5,1), (5,3), (5,5) and suppose one would want to form a region in an L-shape containing (1,1), (1,3), (3,3), then a path to separate them from the rest of the rule cells would for example be (4,0), (4,1), (4,2), (3,2), (2,2), (2,3), (2,4), (1,4), (0,4). The rest of the path cells around them at the edge don't have to be visited since they are only on the outside (x is (0 or width - 1) OR (y is (0 or height - 1)) and therefore not needed to seperate regions.

3. **Path-Based Rules (Edge Touching):**
* **Triangles** (one, two, three or four): The path **MUST** touch a specific number of edges of the cell containing the triangle symbol.
   * One Triangle: Path touches **EXACTLY 1** edge of the triangle's cell.
   * Two Triangle: Path touches **EXACTLY 2** edges of the triangle's cell.
   * Three Triangle: Path touches **EXACTLY 3** edges of the triangle's cell.
   * Four Triangle: Path touches **EXACTLY 4** edges (fully surrounds) the triangle's cell.
* Example: If a cell at (3, 3) has two triangles, the path has to go through **EXACTLY** two places of top (3, 2), right (4, 3), bottom (3, 4), or left (2, 3) to touch the rule cell.

## Task & Output Format
1. **Identifying Objects:** Analyze the grid to identify the coordinates of the Start Node, End Node, and all objects (dots and gaps on path cells; squares, stars, triangles, polyshapes on rule cells - not all types must be present in the puzzle). Keep in mind that rule cells are guaranteed to be located at coordinates where both `x` and `y` are odd. The path MUST NOT pass through these cells.
2. **Solve the Puzzle:** Determine the valid path from the Start Node to the End Node that satisfies all rules. Double check that the path doesn't go through any rule cells (where both coordinates are odd). This is the most common beginner mistake.
3. **Explain Reasoning:** Provide a step-by-step explanation of your thought process. Detail key deductions, how constraints were applied, and any backtracking or choices made.
4. **Provide Solution Path:** After the reasoning, output the exact marker string `####` followed immediately by the solution path as a list of node coordinates `(x, y)`. Include all intermediate nodes from start to end (the distance between each node must be 1).

**Example Solution Path Format:**
####
[(0, 0), (1, 0), (2, 0), (2, 1), ...]
\end{lstlisting}

\subsection{Object-Detection Prompt}
\label{app:prompt-detection}

The object-detection prompt replaces the solving objective with a perception-only objective: rather than producing a path, the model must reconstruct the symbolic grid representation of the puzzle from the rendered image.
The model outputs a 2D array of symbols using the legend defined inside the prompt, which we compare against the ground-truth grid to measure perception accuracy in isolation from path-finding.
The placeholders \texttt{\{start\_marker\}} and \texttt{\{end\_marker\}} follow the same convention as in Appendix~\ref{app:prompt-improved}.

\begin{lstlisting}[basicstyle=\ttfamily\scriptsize, breaklines=true, breakatwhitespace=true, frame=single, columns=fullflexible, keepspaces=true, xleftmargin=0.5em, xrightmargin=0.5em]
You are an expert spatial detection AI specializing in detecting objects from the game 'The Witness'.
Your task is to provide an array of all objects present in the puzzle grid following the specified format.

## Grid and Object Basics
* **Grid Dimensions:** The puzzle grid has {cols} columns and {rows} rows.
* **Coordinate System:** Nodes are identified by `(x, y)` coordinates. `(0,0)` is the top-left node. `({max_x},{max_y})` is the bottom-right node. `x` increases to the right, `y` increases downwards. Nodes can either be path cells or rule cells.
* **Path Cells:** Path cells are dark grey and are at all positions where at least one of `x` or `y` is even. Path cells can contain dots (black hexagons) or gaps (absence of a path where there would normally be one) but can also be empty (normal path).
* **Rule Cells:** Light green cells are rule cells and have coordinates where both `x` and `y` are odd. Rule cells can contain rule symbols (square, star, triangles, polyshapes) but can also be empty.
* **Start and End Nodes:** Exactly one of the path cells on the edge of the board contains a `Start Node` (big circle on the line in the same color as the grey path{start_marker}) and exactly one of the path cells on the edge of the board contains an `End Node` (node from which you can escape the grid to the rounded path outside{end_marker}). Both these nodes are guaranteed to be on valid path cells on the edge of the board (at least one of `x` and `y` has to be an even number).
* **Polyshapes and Negative Polyshapes:** One or multiple filled squares (polyshapes) or unfilled squares (negative polyshapes) in a specific arrangement that defines a shape. The arrangement of the squares defines the shape.
* **Shape Representation:** Represent the shape (Y) of polyshapes and negative polyshapes as a string of 1s and 0s, where 1 represents a filled square and 0 represents an empty square and - represents a line break (from top top left to bottom right). For example, a 2x2 square would be represented as "11-11", a small L-shape could be "10-11", and a small T-shape could be "111-010", a large T-shape could be "111-010-010".
* **Color Codes:** R=Red, B=Blue, G=Green, Y=Yellow, W=White, O=Orange, P=Purple, K=Black

## Symbol Legend
* `S`: **Start Node**
* `E`: **End Node**
* `+`: Empty path cell
* `N`: Empty rule cell
* `G`: **Gap** (gap where a path would normally be)
* `.`: **Dot** (black hexagon on a path cell)
* `o-X`: **Square** of color X (fills our around half of the rule cell)
* `*-X`: **Star** of color X
* `A-X`: **1 Triangle** of color X
* `B-X`: **2 Triangles** of color X
* `C-X`: **3 Triangles** of color X
* `D-X`: **4 Triangles** of color X
* `P-X-Y`: **Polyshape** (positive) of color X and shape Y (one or multiple filled small squares in a specific arrangement)
* `Y-X-Y`: **Negative Polyshape** (ylop) of color X and shape Y (one or multiple unfilled small squares in a specific arrangement)

## Task & Output Format
1. **Identifying Objects:** Analyze the grid to identify the coordinates of all objects and non-objects. Keep in mind that rule cells are guaranteed to be located at coordinates where both `x` and `y` are odd.
2. **Identify Object Colors and Shapes:** For every detected object, determine its color and, for polyshapes and negative polyshapes, its shape. Double check the parity convention: rule cells are at coordinates where both `x` and `y` are odd, and path cells are at coordinates where at least one of `x` and `y` is even. Do NOT solve the puzzle; report only what is visible in the image.
3. **Explain Reasoning:** For all coordinates, write down objects, colors, shapes and explain your reasoning if things were unclear.
4. **Provide Solution Array:** After the reasoning, output the exact marker string `####` followed immediately by the solution array as a list of list of strings (use '' to indicate a string). Include the abbreviation for all objects and all non-objects at all coordinates in the specified format. The output *MUST* follow this format to be correctly parsed.

**Example Solution Format:**
* Assume we have a 5x5 grid with a Start Node at (0, 3) and an End Node at (4, 2) and a black hexagon (dot) at (2, 0) with the rest of the path cells empty and rule cells at (1,1), (1,3), (3,1), (3,3) with a red square at (1,1), 3 green triangles at (1,3) and positive polyshapes (multiple filled squares) of color blue and small L shape at (3,3). Then the output should end with:
####
[['+', '+', '+', 'S', '+'], ['+', 'o-R', '+', 'C-G', '+'], ['.', '+', '+', '+', '+'], ['+', 'N', '+', 'P-B-11-10', '+'], ['+', '+', 'E', '+', '+']]
\end{lstlisting}

%% file: annotation.tex
\clearpage
\onecolumn
\hypertarget{annotation}{}
\pagestyle{empty}
\lstset{
  basicstyle=\footnotesize\ttfamily,
  breaklines=true,
  breakatwhitespace=false,
  columns=flexible,
  numbers=none
}

\definecolor{Primary}{RGB}{59, 130, 246}    %
\definecolor{PrimaryDark}{RGB}{30, 64, 175} %
\definecolor{LightBg}{RGB}{239, 246, 255}   %
\definecolor{TextDark}{RGB}{31, 41, 55}     %
\definecolor{TextMuted}{RGB}{107, 114, 128} %

\begin{tikzpicture}[remember picture, overlay]
  \fill[Primary] ([xshift=0cm,yshift=0cm]current page.north west) rectangle ([xshift=\paperwidth,yshift=-0.4cm]current page.north west);
\end{tikzpicture}

\vspace{0.8cm}
\begin{center}
  {\fontsize{22}{26}\selectfont\sffamily\bfseries \textcolor{PrimaryDark}{CiteAssist}}\\[0.2em]
  {\Large\sffamily\scshape \textcolor{TextMuted}{Citation Sheet}}\\[0.8em]
  {\small\sffamily Generated with \href{https://citeassist.uni-goettingen.de/}{\textcolor{Primary}{\texttt{citeassist.uni-goettingen.de}}}
  \CiteAssistCite{}
  }\end{center}

\begin{center}
\vspace{1em}
\begin{tikzpicture}
\draw[Primary, line width=0.6pt] (0,0) -- (\textwidth,0);
\end{tikzpicture}
\vspace{1.2em}
\end{center}

\begin{tcolorbox}[enhanced,
                 frame hidden,
                 boxrule=0pt,
                 borderline west={2pt}{0pt}{Primary},
                 colback=LightBg,
                 sharp corners,
                 breakable,
                 fonttitle=\sffamily\bfseries\large,
                 coltitle=Primary,
                 title=BibTeX Entry,
                 attach title to upper={\vspace{0.2em}\par},
                 left=12pt]
\lstset{
    inputencoding = utf8,  %
    extendedchars = true,  %
    literate      =        %
      {á}{{\'a}}1  {é}{{\'e}}1  {í}{{\'i}}1 {ó}{{\'o}}1  {ú}{{\'u}}1
      {Á}{{\'A}}1  {É}{{\'E}}1  {Í}{{\'I}}1 {Ó}{{\'O}}1  {Ú}{{\'U}}1
      {à}{{\`a}}1  {è}{{\`e}}1  {ì}{{\`i}}1 {ò}{{\`o}}1  {ù}{{\`u}}1
      {À}{{\`A}}1  {È}{{\`E}}1  {Ì}{{\`I}}1 {Ò}{{\`O}}1  {Ù}{{\`U}}1
      {ä}{{\"a}}1  {ë}{{\"e}}1  {ï}{{\"i}}1 {ö}{{\"o}}1  {ü}{{\"u}}1
      {Ä}{{\"A}}1  {Ë}{{\"E}}1  {Ï}{{\"I}}1 {Ö}{{\"O}}1  {Ü}{{\"U}}1
      {â}{{\^a}}1  {ê}{{\^e}}1  {î}{{\^i}}1 {ô}{{\^o}}1  {û}{{\^u}}1
      {Â}{{\^A}}1  {Ê}{{\^E}}1  {Î}{{\^I}}1 {Ô}{{\^O}}1  {Û}{{\^U}}1
      {œ}{{\oe}}1  {Œ}{{\OE}}1  {æ}{{\ae}}1 {Æ}{{\AE}}1  {ß}{{\ss}}1
      {ẞ}{{\SS}}1  {ç}{{\c{c}}}1 {Ç}{{\c{C}}}1 {ø}{{\o}}1  {Ø}{{\O}}1
      {å}{{\aa}}1  {Å}{{\AA}}1  {ã}{{\~a}}1  {õ}{{\~o}}1 {Ã}{{\~A}}1
      {Õ}{{\~O}}1  {ñ}{{\~n}}1  {Ñ}{{\~N}}1  {¿}{{?\`}}1  {¡}{{!\`}}1
      {„}{\quotedblbase}1 {“}{\textquotedblleft}1 {–}{$-$}1
      {°}{{\textdegree}}1 {º}{{\textordmasculine}}1 {ª}{{\textordfeminine}}1
      {£}{{\pounds}}1  {©}{{\copyright}}1  {®}{{\textregistered}}1
      {«}{{\guillemotleft}}1  {»}{{\guillemotright}}1  {Ð}{{\DH}}1  {ð}{{\dh}}1
      {Ý}{{\'Y}}1    {ý}{{\'y}}1    {Þ}{{\TH}}1    {þ}{{\th}}1    {Ă}{{\u{A}}}1
      {ă}{{\u{a}}}1  {Ą}{{\k{A}}}1  {ą}{{\k{a}}}1  {Ć}{{\'C}}1    {ć}{{\'c}}1
      {Č}{{\v{C}}}1  {č}{{\v{c}}}1  {Ď}{{\v{D}}}1  {ď}{{\v{d}}}1  {Đ}{{\DJ}}1
      {đ}{{\dj}}1    {Ė}{{\.{E}}}1  {ė}{{\.{e}}}1  {Ę}{{\k{E}}}1  {ę}{{\k{e}}}1
      {Ě}{{\v{E}}}1  {ě}{{\v{e}}}1  {Ğ}{{\u{G}}}1  {ğ}{{\u{g}}}1  {Ĩ}{{\~I}}1
      {ĩ}{{\~\i}}1   {Į}{{\k{I}}}1  {į}{{\k{i}}}1  {İ}{{\.{I}}}1  {ı}{{\i}}1
      {Ĺ}{{\'L}}1    {ĺ}{{\'l}}1    {Ľ}{{\v{L}}}1  {ľ}{{\v{l}}}1  {Ł}{{\L{}}}1
      {ł}{{\l{}}}1   {Ń}{{\'N}}1    {ń}{{\'n}}1    {Ň}{{\v{N}}}1  {ň}{{\v{n}}}1
      {Ő}{{\H{O}}}1  {ő}{{\H{o}}}1  {Ŕ}{{\'{R}}}1  {ŕ}{{\'{r}}}1  {Ř}{{\v{R}}}1
      {ř}{{\v{r}}}1  {Ś}{{\'S}}1    {ś}{{\'s}}1    {Ş}{{\c{S}}}1  {ş}{{\c{s}}}1
      {Š}{{\v{S}}}1  {š}{{\v{s}}}1  {Ť}{{\v{T}}}1  {ť}{{\v{t}}}1  {Ũ}{{\~U}}1
      {ũ}{{\~u}}1    {Ū}{{\={U}}}1  {ū}{{\={u}}}1  {Ů}{{\r{U}}}1  {ů}{{\r{u}}}1
      {Ű}{{\H{U}}}1  {ű}{{\H{u}}}1  {Ų}{{\k{U}}}1  {ų}{{\k{u}}}1  {Ź}{{\'Z}}1
      {ź}{{\'z}}1    {Ż}{{\.Z}}1    {ż}{{\.z}}1    {Ž}{{\v{Z}}}1  {ž}{{\v{z}}}1
  }
\begin{lstlisting}
@inproceedings{kaesberg2026,
  author={Kaesberg, Lars Benedikt and Yang, Tianyu and Wunderlich, Florian Valentin
and Ruas, Terry and Kurzawe, Daniel and Wahle, Jan Philip and Gipp, Bela},
  booktitle={Findings of the Association for Computational Linguistics: {EMNLP} 2026},
  publisher={Association for Computational Linguistics},
  title={Is Visual Prompting All You Need? Studying {VLM} Spatial Reasoning under
Progressive Visual Scaffolds},
  topic={nlp},
  year={2026}
}
\end{lstlisting}
\end{tcolorbox}

\vfill
\noindent\begin{tikzpicture}
\draw[Primary!40, line width=0.4pt] (0,0) -- (\textwidth,0);
\end{tikzpicture}
\begin{center}
\small\sffamily\textcolor{TextMuted}{Generated \today}
\end{center}